\documentclass[10pt,journal]{IEEEtran}

\ifCLASSINFOpdf
\else
\fi
\usepackage{times}
\usepackage{epsfig}
\usepackage{graphicx}
\usepackage{amsmath}
\usepackage{amssymb}

\usepackage{footnote}
\usepackage{multirow}
\usepackage{booktabs}
\usepackage{threeparttable}
\usepackage{enumerate}
\usepackage{indentfirst}
\usepackage{array}
\usepackage{rotating}
\usepackage{ragged2e}
\usepackage{wasysym}
\usepackage{caption}
\usepackage{epstopdf}
\usepackage{color}
\usepackage{algorithm}
\usepackage{times}
\usepackage{epsfig}
\usepackage{graphicx}
\usepackage{amsmath}
\usepackage{amssymb}
\usepackage{epstopdf}
\usepackage{color}
\usepackage{array}
\usepackage{tabu} 
\usepackage{comment}
\usepackage{algorithm}
\usepackage{algorithmicx}
\usepackage{cases}
\usepackage{bm}
\usepackage{bbm}
\usepackage{algpseudocode}
\usepackage{booktabs}
\usepackage{multirow}
\usepackage{textcomp}

\def\ie{{\em i.e.}}
\def\eg{{\em e.g.}}
\def\etal{{\em et al.}}

\newcommand{\secref}[1]{Section \ref{#1}}

\newcommand{\br}[1]{\bm{\mathrm{#1}}}

\begin{document}
	%
	\title{Polarization Guided Specular Reflection Separation}
	%
	%
	%
	
	\author{Sijia~Wen,
		Yinqiang~Zheng,~\IEEEmembership{Senior Member,~IEEE}
		and~Feng~Lu,~\IEEEmembership{Member,~IEEE}
		\IEEEcompsocitemizethanks{
			\IEEEcompsocthanksitem Manuscript received June 08, 2020; revised December 09, 2020 and June 28, 2021; accepted August 02, 2021. This work was supported by National Natural Science Foundation of China (NSFC) under Grants 61972012 and 61732016. (\emph{Corresponding author: Feng Lu})\protect
			\IEEEcompsocthanksitem  Sijia Wen and Feng Lu are with the State Key Laboratory of Virtual Reality Technolog  and Systems, School of Computer Science and Engineering, Beihang University, Beijing 100191, China, and also with the Peng Cheng Laboratory, Shenzhen 518066, China. (e-mail: sijiawen@buaa.edu.cn; lufeng@buaa.edu.cn)\protect
			\IEEEcompsocthanksitem Y. Zheng is with the Next Generation Artificial Intelligence Research Center, The University of Tokyo, Tokyo 113-8656, Japan. (e-mail: yqzheng@ai.u-tokyo.ac.jp). \protect}
	}
	
	%
	%

	\markboth{Submission to IEEE TRANSACTIONS ON IMAGE PROCESSING}%
	{Shell \MakeLowercase{\textit{et al.}}: Bare Demo of IEEEtran.cls for IEEE Journals}
	%



	\maketitle
	
	\begin{abstract}
		Since specular reflection often exists in the real captured images and causes deviation between the recorded color and intrinsic color, specular reflection separation can bring advantages to multiple applications that require consistent object surface appearance. However, due to the color of an object is significantly influenced by the color of the illumination, the existing researches still suffer from the near-duplicate challenge, that is, the separation becomes unstable when the illumination color is close to the surface color. In this paper, we derive a polarization guided model to incorporate the polarization information into a designed iteration optimization separation strategy to separate the specular reflection. Based on the analysis of polarization, we propose a polarization guided model to generate a polarization chromaticity image, which is able to reveal the geometrical profile of the input image in complex scenarios, \eg, diversity of illumination. The polarization chromaticity image can accurately cluster the pixels with similar diffuse color. We further use the specular separation of all these clusters as an implicit prior to ensure that the diffuse component will not be mistakenly separated as the specular component. With the polarization guided model, we reformulate the specular reflection separation into a unified optimization function which can be solved by the ADMM strategy. The specular reflection will be detected and separated jointly by RGB and polarimetric information. Both qualitative and quantitative experimental results have shown that our method can faithfully separate the specular reflection, especially in some challenging scenarios.
	\end{abstract}
	

	%
	\IEEEpeerreviewmaketitle

	\section{Introduction}
	
	\IEEEPARstart{T}{he} reflection image from the surface consists of the diffuse component and the specular component. They are often observable in dielectric inhomogeneous objects~\cite{shafer1985using}. Lee~\etal~\cite{lee1990modeling} propose that the specular reflection is significantly affected by the illumination of the scene while the diffuse reflection contains the constant intrinsic properties of the surface. 
	
	For multiple computer vision tasks, such as segmentation~\cite{dai2016instance}, shape from shading~\cite{zhu2019depth,chen2020learning}, binocular stereo~\cite{cui2017polarimetric}, and motion detection~\cite{singla2014motion}, the specular component is often considered as the outliers. Since the specular component is mostly combined with the diffuse component at each pixel in the image~\cite{tominaga1991surface,shafer1985using}, the above-mentioned tasks have to ignore the specular component in the scene. However, assuming the pure-diffuse surface decreases the performance of those applications. Besides, the specular removal can also help to resolve the general reflectance problem~\cite{lu2015intensity,lu2015uncalibrated} in photometric stereo. To address this challenging issue, various existing methods have been proposed on the benefit of chromaticity or polarization.
	\begin{figure}[t]
		\begin{center}
			\includegraphics[width=0.95\linewidth]{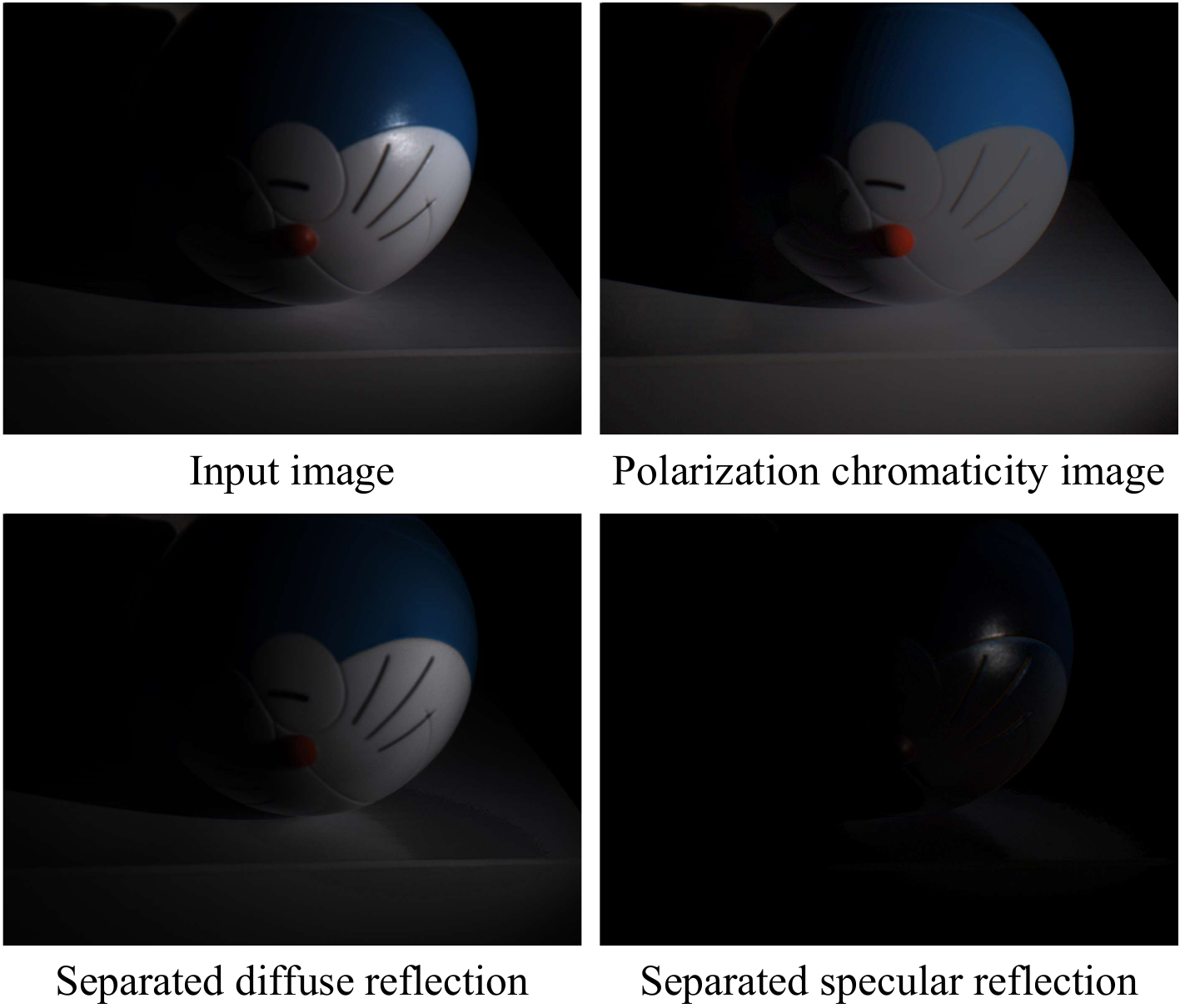}
		\end{center}
		\caption{An exemplar of a scene, where the color of the object in the scene is similar to the color of illumination.}
		\label{ex}
	\end{figure}
	
	Based on the analysis of color space and spatial information, several methods~\cite{klinker1990physical,yang2012new,yang2013separating,yang2014efficient} with some certain prior knowledge have been proposed to remove the specular reflection by understanding the distribution of diffuse component and specular component in a color image. A typical way is to generate the specularity-free image~\cite{akashi2014separation,bajcsy1996detection,shen2008chromaticity,article}. These methods propagate the chromaticity from the specularity-free image to the original image. According to the dichromatic reflection model, other methods~\cite{yang2010real,akashi2014separation} are obliged to assume or estimate the illumination of the scene. However, in the real world, illumination is inconstant and hard to measure. In addition, since these algorithms rely on either image statistics or strong prior assumptions, they are not robust for the variability in the imaging environment.
	
	Based on the theory of polarization~\cite{huard1997polarization,wolff1993constraining}, the specular reflection tends to be polarized meanwhile the diffuse reflection is unpolarized~\cite{born2013principles}. By placing the polarization filter in front of the sensor, various methods~\cite{nayar1997separation,umeyama2004separation,wang2017specularity,guo2018single} can produce the results with less color distortion. However, since the correlation between RGB information and polarimetric information was not fully exploited, these methods cannot obtain pleasing results. In this regard, considering the complexity and uncertainty among different illumination environment, specular reflection separation is still a very challenging task.
	
	In this paper, we propose a polarization guided specular reflection separation approach to achieve wide applicability for the large diversity of nature scenes. According to the theory of polarization, the diffuse component tends to be unpolarized while the specular component varies with the different angles of polarization orientation. Motivated by this observation, we can obtain the raw diffuse image by the transmitted radiance sinusoid (TRS). Since the raw diffuse image still consists of the diffuse component and the constant part of the specular component, it cannot be considered as the result of specular reflection separation. However, by calculating the chromaticity of the raw diffuse image, we can generate a \br{\textit{polarization chromaticity image}}, while retaining the hue of it. The polarization chromaticity image is able to describe the accurate geometrical profile of the input image, without being affected by the color of the illumination. With the polarization chromaticity image, the polarization guided model can cluster the pixels with similar diffuse color. In this case, the specular component can be considered as the noise in each cluster. After denoising for all these clusters by using robust PCA~\cite{candes2011robust}, most parts of the specular component can be removed from the observation. 
	
	Even though the polarization guided model can deliver a pleasing result, the specular reflection is partially polarized, which means we still need to estimate the unchanged part of the specular component. Different from setting some prior knowledge, we use the result of the polarization guided model as an implicit prior to the diffuse component and impose the sparse constraint on the specular component. With the proposed model, we reformulate the specular reflection separation to a global energy function that can be optimized by the ADMM strategy. Moreover, we collect a dataset by the newly released polarized RGB camera. The corresponding ground truth is captured by rotating the polarizer in front of an RGB camera. Experimental results demonstrate the effectiveness and robustness of our method.
	
	Our main contributions are summarized as follows: 1) We design a polarization guided model for specular reflection separation without being affected by the color of the illumination; 2) We design a customized optimization strategy to achieve the promising results jointly by chromaticity and polarization; 3) We conduct extensive experiments to demonstrate that our proposed method achieves state-of-the-art results in terms of quantitative measures and visual quality.
	
	The rest of this paper is organized as follows: \secref{related work} reviews related work and \secref{PGM} details the polarization guided model. \secref{Approach} presents the scheme of the specular reflection separation and \secref{Ex} describes the experimental results. Finally, \secref{conclusion} concludes the paper.
	
	\section{Related Work}\label{related work}
	
	Since specular reflection separation has been attracting increasing interest in computer vision, plenty of researches have been presented on either image quality improvement or applying it to specific applications~\cite{khan2017analytical,artusi2011survey}. The existing methods for the specular reflection separation can be generally grouped into two categories. The first one uses a single image as the input. Due to the fact that the specular reflection changes significantly in different observations, the other one uses a sequence of images as the input. In this section, we overview all these categories as related works. 
	
	\subsection{Single-image Methods}
	
	The chromatic information can help to analyze the distribution of diffuse and specular pixels in the RGB space. Klinker~\etal~\cite{klinker1987using,klinker1988measurement} classify color pixels in the categories of diffuse, specular, and saturated pixels. Based on the analysis of the color histogram and the convex polygon fitting technique, they separate reflection components by fitting them into a dichromatic plane. To fully exploit the color space of the image, transferring from the RGB space to another imagery space is also an effective method. Schl{\"u}ns~\etal~\cite{schluns1995analysis,schluns1995fast} transform the image to YUV color space. Bajcsy~\etal~\cite{bajcsy1996detection} propose the S-space for analysis of variation in color of objects. Based on the Ch-CV space~\cite{yang2012new}, Yang~\etal~\cite{yang2013separating}
	propose a separation method in HSI color space. Tan~\etal~\cite{tan2006separation} present a technique that makes use of texture data to overcome the typical problems from color space analysis methods.
	
	In order to improve the robustness and effectiveness of the separation results, Tan~\etal~\cite{article}, Yoon~\etal~\cite{yoon2006fast} and Shen~\etal~\cite{shen2009simple} utilize the specularity-free image to remove the specular component. Yang~\etal~\cite{yang2010real} propose a simple and effective method by directly applying the low-pass filter to the maximum fraction of the color component of the original image.  Mallik~\etal~\cite{mallick2006specularity} introduce a partial differential equation (PDE) that iteratively erodes the specular component at each pixel. Tappen~\etal~\cite{tappen2003recovering} present a method that uses color information and image derivative classifiers to recover the diffuse and specular intrinsic properties of an image. These pixel-wise methods can deliver more pleasing results, yet at the cost of heavy computation.
	
	By using the optimization function with the constraint of some prior, these works~\cite{akashi2014separation,suo2016fast,liu2015saturation,ren2017specular,guo2018single} can reduce the computational cost. However, all of them are based on spatial prior knowledge that is only applicable to general conditions. It will reduce the robustness of the algorithm. In addition, the RGB information cannot accurately separate the specular reflection in complex scenarios, such as under different colors of the illumination. 
	
	\subsection{Multi-images Methods}
	
	Instead of analyzing the chromatic information in a single image, the multi-image methods use the information contained in an image sequence, which is from different points of view or with different light information. Since the information contained in such an image sequence is richer than in a single image, the multi-images methods can deliver better separation results.
	
	\subsubsection{Color-based methods}
	\begin{figure*}[t]	
		\centering
		\includegraphics[width=1\linewidth]{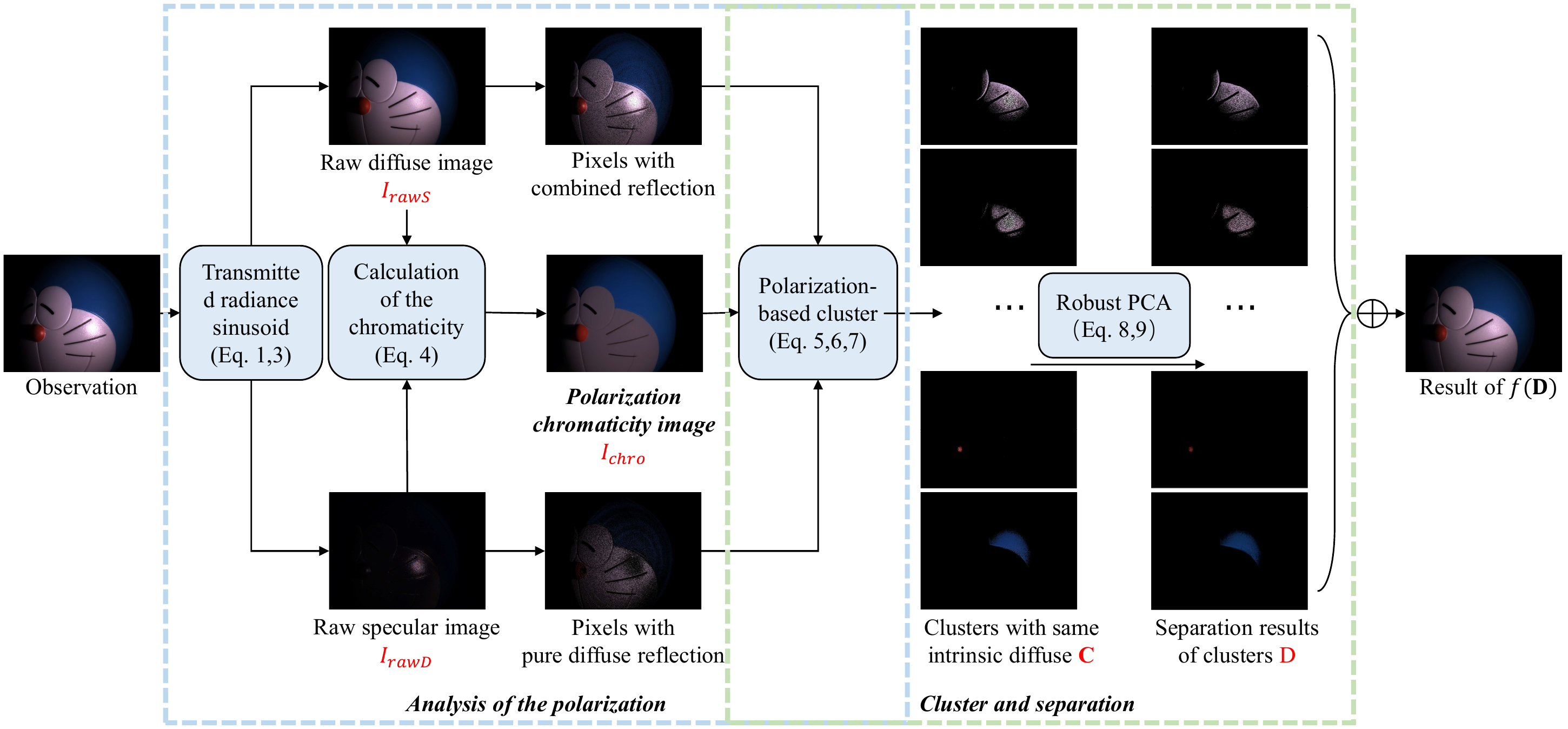}
		\caption{The pipeline of the proposed polarization guided model.}
		\label{pipeline}
	\end{figure*}
	
	Based on the assumption of Lambertian consistency, early work proposed by Lee~\etal~\cite{lee1992detection} uses images from different points of view to locate the specular area. Later, several works~\cite{sato1994temporal,lin2001separation,chen2006mesostructure} set different illumination for the same scene to remove the specular component. On the basis of multi-view stereo, Lin~\etal~\cite{lin2002diffuse} and Yang~\etal~\cite{yang2010uniform} use stereo images to detect the specular component. Weiss~\etal~\cite{weiss2001deriving} obtain the intrinsic images of the scene by assuming that the sparsity of specular component various with different illumination. Agrawal~\etal~\cite{agrawal2005removing} propose a method for image enhancement by using two images of a scene. 
	
	All these color-based methods rely on the observation that the specular component will be various under different viewing angles. The location of the specular area is also affected by either intensity of illumination or angles of the incident light. However, in the practical scene, part of the specular component is always constant. It will decrease the performance of these above-mentioned methods.
	
	\subsubsection{Polarization-based methods}
	Different from the color-based methods, to avoid the color distortion caused by illumination, polarization-based methods take advantage of polarimetric information. In general condition, the diffuse reflection is unpolarized while the specular reflection is polarized, which can be considered as a strong indicator of the specular component.
	
	Nayar~\etal~\cite{nayar1997separation} present a method by analyzing the scene through the direct and global reflection components, which have shown good performance in their applications. Yet the pixel-wise calculation limits the effectiveness of performance. By estimating the fixed coefficients of the specular component, Umeyama~\etal~\cite{umeyama2004separation} design a global polarization-based algorithm which can be solved by Independent Component Analysis (ICA)~\cite{hyvarinen2000independent}. Based on the approach of Umeyama, Wang~\etal~\cite{wang2017specularity} replace the fixed coefficient with the spatially variable coefficient to improve the performance of the specular reflection separation. However, all these works still require a strict controllable light source, which limits the applicability of the performance. Using images acquired with three different angles of polarization, Zhang~\etal~\cite{zhang2011reflection} directly separate the reflection by TRS. However, since the specular reflection is partially polarized, the separation in this way still retains some parts of the specular component in the result.
	
	Therefore, in spite of the fact that numerous researches have been conducted for specular removal, there are still some shortcomings and limitations in the specular reflection separation. In this regard, we propose a polarization guided method to separate the specular reflection jointly by chromaticity and polarization.
	
	\section{Polarization Guided Model }\label{PGM}
	In this paper, we propose the polarization guided model, which aims to separate the specular reflection under complex scenarios. Since the result of the polarization guided model is imposed on the processing of specular reflection separation as a constraint, whether the polarization guided model can portray the characteristics of the specular reflection separation will play a key role in the final result. Based on the analysis of polarization, the proposed model generates a polarization chromaticity image, which is able to reveal the geometrical profile of the input image in complex scenarios, such as diversity of illumination. With the polarization chromaticity image, the observation can be divided into clusters, each of which contains the pixels with a similar intrinsic diffuse color. Inspired by the Robust PCA~\cite{candes2011robust}, we remove the specular component from each cluster. The pipeline of the polarization guided model is illustrated in Fig.~\ref{pipeline}.
	\begin{figure}[htbp]	
		\centering
		\includegraphics[width=1\linewidth]{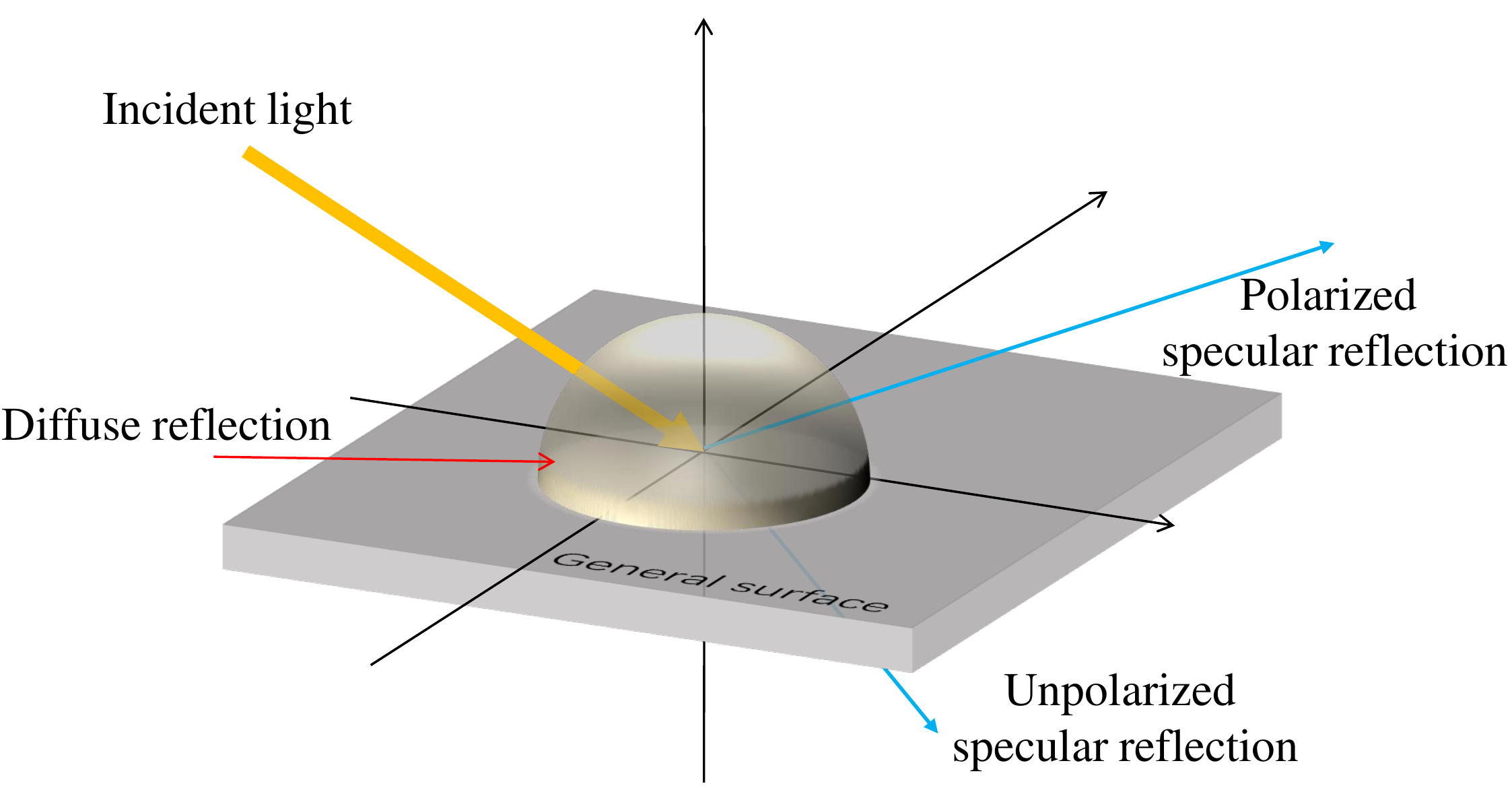}
		\caption{The reflected light is generally partially polarized in the real scene. The complete linear polarization only exists, if the light strikes the interface at Brewster's angle~\cite{brewster1815laws}.}
		\label{light}
	\end{figure}
	
	\subsection{Analysis of the Polarization }\label{ap}
	According to Fresnel’s theory~\cite{winthrop1965theory}, the diffuse component $I_{d}$ maintains constant while the specular component $I_{s}$ varies under different angles of polarization orientation. Since the specular reflection is partially polarized, part of the specular component $I_{s}$ is also constant, as shown in Fig.~\ref{light}. 
	
\begin{figure}[b]
		\centering
		\includegraphics[width=0.78\linewidth]{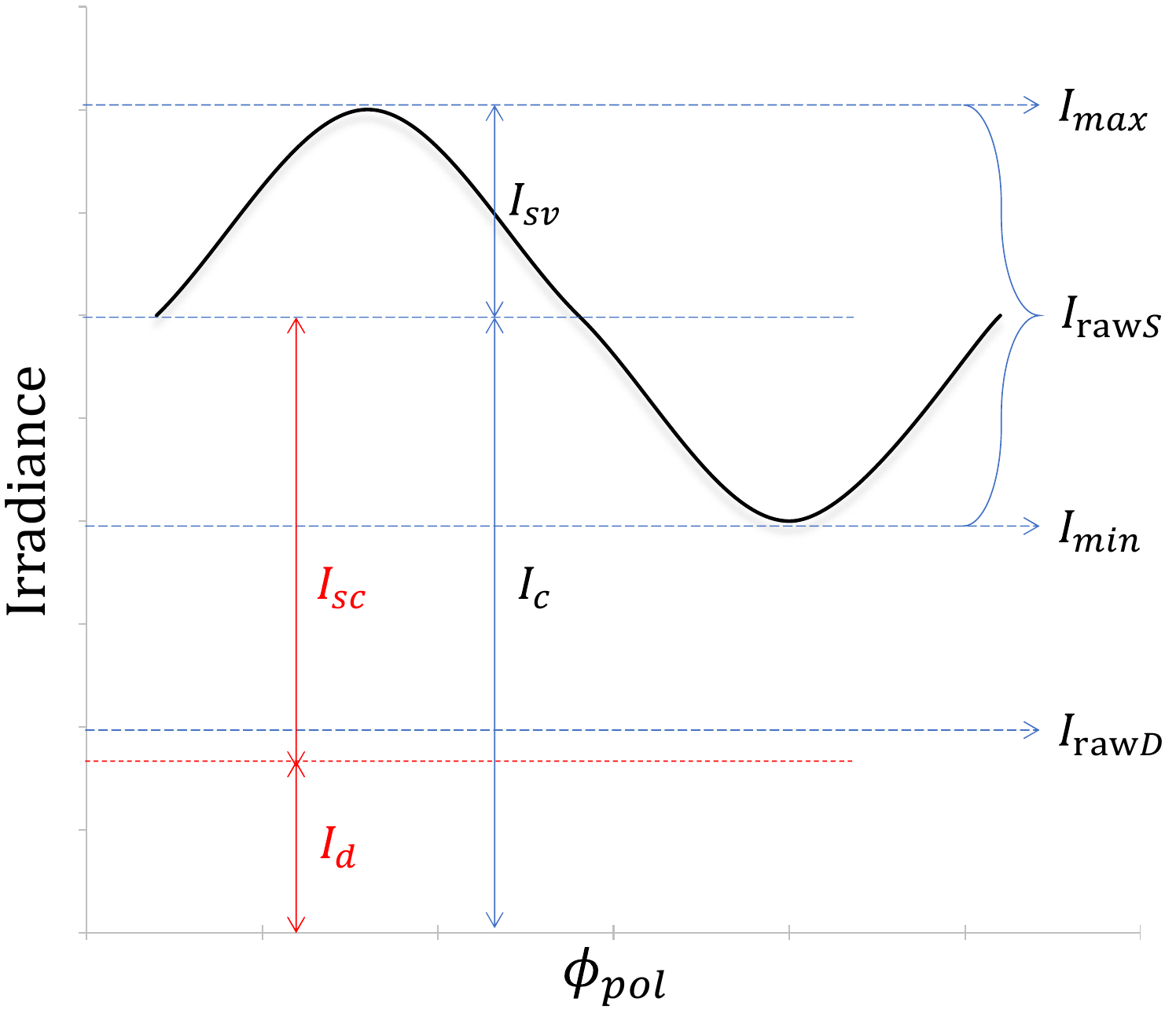}
		\caption{Image brightness plotted as a function of polarization filter angle.}
		\label{curve}
	\end{figure}

	Therefore, with different polarization orientation $\phi_{pol}$, the specular component can be expressed as the sum of a constant component $I_{sc}$ and a cosine function term with amplitude $I_{sv}$. As shown in Fig.~\ref{curve}, by plotting the intensity of a pixel across the different angle of polarization orientation $\phi_{pol}$, the various value of irradiance is following by:
	\begin{equation} \label{E4}
	\begin{split}
	{I(\phi_{pol})} &= I_d + I_{sc} + I_{sv}\cos2(\phi_{pol} - \alpha) \\
	&= I_c + I_{sv}\cos2(\phi_{pol} - \alpha),
	\end{split}
	\end{equation}
	where $I_c$ is the constant reflection, which is the sum of diffuse component $I_d$ and unpolarized specular component $I_{sc}$. $\alpha$ is the phase angle (\ie, the angle of the polarization axis, relative to a global coordinate system).
	
	With the latest progress in image sensor manufacturing technology, it is now possible to simultaneously capture RGB and polarimetric information of the scene $\left\{I(\phi_{pol}) | \phi_{pol}\in(0^\circ,45^\circ,90^\circ,135^\circ)\right\}$. Eq.~\ref{E4} can be written in the following vector notation:
	\begin{equation} \label{E5}
	\begin{aligned}
	&P = (1, \cos2\phi_{pol}, \sin2\phi_{pol}),\\
	&O = (I_{c}, I_{sv}\cos2\alpha, I_{sv}\sin2\alpha),\\
	&I(\phi_{pol}) = PO,
	\end{aligned}
	\end{equation}
	where $P$ is given and $I(\phi_{pol})$ is the observation. We can calculate $I_c$ and $I_{sv}$ by solving Eq.~\ref{E5} as an over-determined linear system of equations~\cite{tibshirani1996regression}. However, separating $I_d$ from $I_c$ is still a challenging problem, as shown in Fig.~\ref{curve}. In this regard, we use $I_c$ and $I_{sv}$ to generate a polarization chromaticity image to address this issue.
	
	\subsection{Polarization Chromaticity Image }\label{PCI}
	\begin{figure*}[htbp]	
		\centering
		\includegraphics[width=0.91\linewidth]{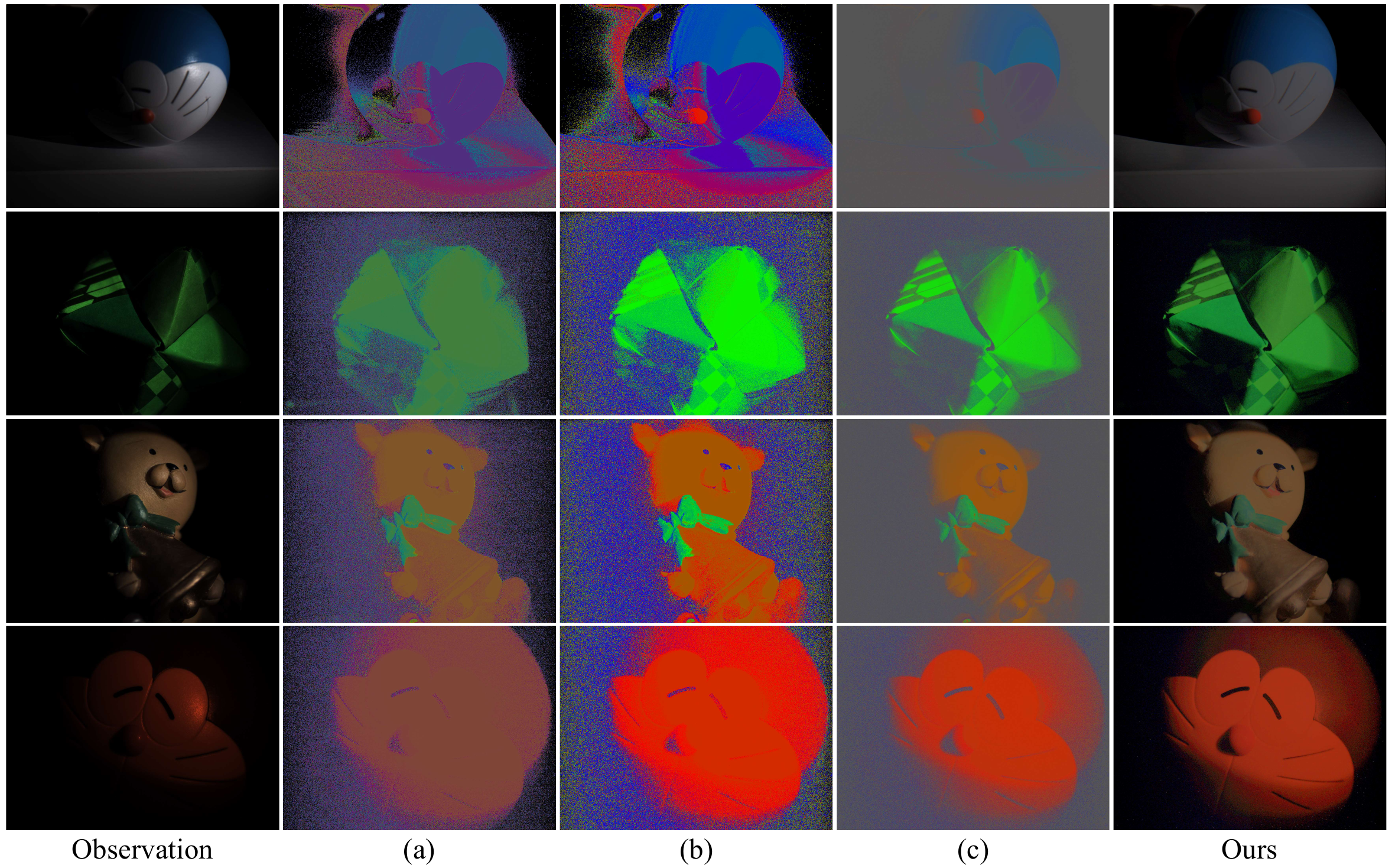}
		\caption{The chromaticity images are generated: (a) by Tan's method~\cite{article}, (b) by Kim's method~\cite{kim2013specular}, (c) by Shen's method~\cite{shen2008chromaticity}, and ours.}
		\label{ChromaticityIamge}
	\end{figure*}
	
	Various researches~\cite{article,shen2008chromaticity,kim2013specular} work on the specular reflection separation with the chromaticity image generated by the specularity-free image. However, when the color of the scene is similar to the illumination, these methods still suffer from the color distortion of the generated chromaticity image. Moreover, all these methods assume the intensity of illumination as [1,1,1]. To generate a robust and accurate chromaticity image, we introduce the polarization to our work. According to \secref{ap}, the variable irradiance definitely belongs to the specular reflection, no matter what color the illumination is. 
	
	Though we cannot directly obtain the diffuse component $I_d$, introducing the approximate diffuse component (raw diffuse image $I_{rawD}$) and approximate specular component (raw specular image $I_{rawS}$) can help for subsequent calculations. Since the variable part of the irradiance should definitely belong to the specular reflection, we directly set this variable component $2*I_{sv}$ as the raw specular image $I_{rawS}$. For $I_{rawD}$, it should be noted that utilizing the input image to calculate $I_{rawD}$ will remain the unpolarized specular component $I_{sc}$, which is verified by the previous works~\cite{nayar1997separation,zhang2011reflection}. In order to obtain the $I_{rawD}$ closer to $I_d$, we remove the raw specular image $I_{rawS}$ from the constant reflection $I_{c}$ as following:
	\begin{equation} \label{E6}
	\begin{aligned}
	&I_{rawS} = 2*I_{sv},\\
	&I_{rawD} = I_c - I_{rawS}.
	\end{aligned}
	\end{equation}
	Since the main purpose of the proposed $I_{rawD}$ is to calculate the polarization chromaticity image, we only require the approximate value of it. In addition, as the initialization, $I_{rawS}$ and $I_{rawD}$ will be optimized by the further proposed algorithm.
	
	Although the raw diffuse image $I_{rawD}$ is different from the pure diffuse image $I_{d}$, $I_{rawD}$ can help to obtain the polarization chromaticity image $I_{chro}$. Regardless of the illumination and the color of the specular component, $I_{chro}$ can reveal the intrinsic diffuse reflection of the image. The calculation of $I_{chro}$ is given by: 
	\begin{equation} \label{E7}
	\begin{aligned}
	&I_{chro} = \frac{I_{rawD}}{\sum_{\theta} I_{rawD,\theta} + \overline{I_{min}}},\\
	&\overline{I_{min}} = \frac{\sum_{p}min(I_{r}(p), I_{g}(p), I_{b}(p))}{N},
	\end{aligned}
	\end{equation}
	where $\theta\in{\{R,G,B\}}$ and $N$ is total number of pixels in the input image. The $p$ in Eq~\ref{E7} stands for each pixel. $\overline{I_{min}}$ is the average of the minimum values in the r, g, b channels of all pixels. The introduction of $\overline{I_{min}}$ will address the unstable situation caused by the dark or noise pixels in the polarization chromaticity image. 
	
	We compare the polarization chromaticity images with the chromaticity images generated by former methods, as shown in Fig.~\ref{ChromaticityIamge}. The comparison results demonstrate that the polarization guided model can deliver a promising chromaticity image, which can accurately describe the scene. As expected, the polarization chromaticity image is insensitive to the noise. In addition, $I_{chro}$ will not be affected when the diffuse color is similar to the illumination. Therefore, the polarization chromaticity image can help to cluster the pixels with a similar intrinsic diffuse color.
	
	\subsection{Cluster and Separation }
	
	Based on the analysis of the polarization, as shown in Fig.~\ref{curve}, the value of $I_{d}$ is definitely smaller than the value of $I_{min}$, which is defined as:
	\begin{equation}\label{E50}
	I_{d}(p) <I_{c}(p) - I_{sv}(p).
	\end{equation}
	According to Eq.~\ref{E6}, Eq.~\ref{E50} can be transformed to:
	\begin{equation}
	I_{d}(p) - I_{rawD}(p) < I_{sv}(p).
	\end{equation}
	As a result, if the different intensity between the pixel $p$ in observation and the one in the raw diffuse image $I_{rawD}$ is smaller than $I_{sv}$, we assume $p$ as a pixel with the pure diffuse reflection. Otherwise, the pixel $q$ will belong to the pixels with combined reflection. Then we can cluster the pixels with a similar intrinsic diffuse color, which is guided by the polarization chromaticity image. As we explained in Sec.~\ref{PCI}, the chromaticity image can sharply describe the scene without being affected by the color of illumination. The cluster will be simply following:
	\begin{equation} \label{E9}
	\begin{aligned}
	I_{chro}(p) - I_{chro}(q) < T,
	\end{aligned}
	\end{equation}
	where $T$ is the chromatic threshold. 
	\begin{figure*}[htbp]	
		\centering
		\includegraphics[width=0.93\linewidth]{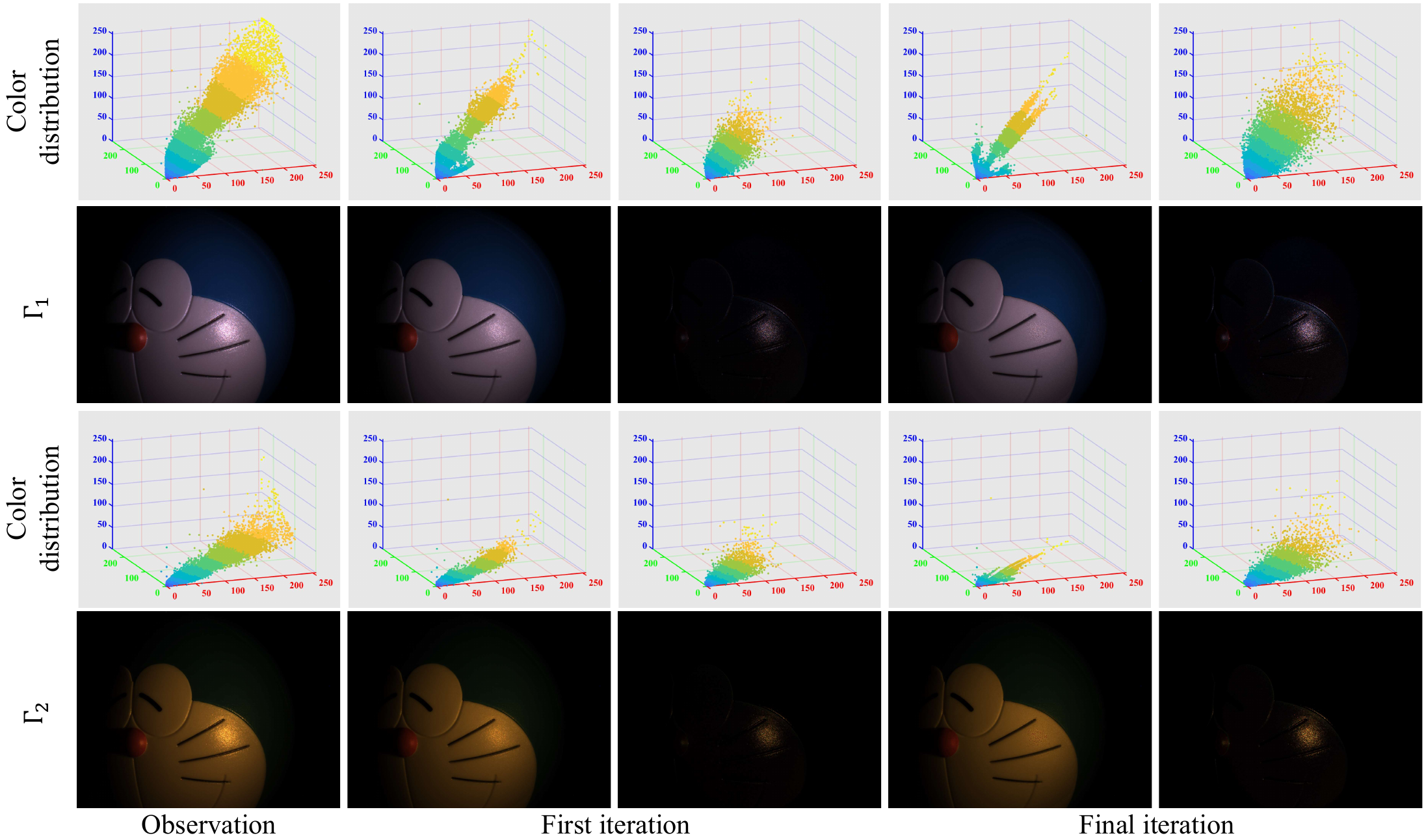}
		\caption{The specular reflection separation results of the proposed method. The first and third rows show the color distribution of pixels in the RGB space. The second and fourth rows are the images captured under different illumination $\Gamma_1$ and $\Gamma_2$. }
		\label{iteration}
	\end{figure*}

	Then we will obtain a cluster $\br{C}$ which contains the diffuse component $\br{D}$ and the specular component $\br{S}$. In a physical sense, the rank of a matrix measures the correlation between the rows and columns of the matrix. Since the diffuse component $\br{D}$ owns a similar intrinsic diffuse color, all the rows of the matrix can be transformed into the clustered intrinsic diffuse color. It explains that the $\br{D}$ contains the main structure of the cluster, which is low-rank. Often, we observe that highlight regions are small in size and are distributed rather sparsely. Therefore, $\br{S}$ can be considered as the noise, which is salient and tends to be sparse. The specular reflection separation of the cluster can further be formulated as the following optimization problem:
	\begin{equation} \label{E10}
	\begin{aligned}
	&\mathop{min} \limits_{\br{D},\br{S}} rank(\br{D}) + \lambda \|\br{S}\|_0\\
	&s.t. \br{C} = \br{D} + \br{S}, \br{D}>0,
	\end{aligned} 	
	\end{equation}
	where $\br{C}$, $\br{D}$ and $\br{S}$ are $\br{X} \times 3$ matrices with each row representing a pixel color. $\br{X}$ is the total number of pixels in the cluster. $\|\cdot \|_0$ denotes the $l_0$ norm of a matrix. To make the optimization tractable and convex, we relax Eq.~\ref{E10} by adding the regularization term:
	\begin{equation} \label{E11}
	\begin{aligned}
	&\mathop{min} \limits_{\br{D},\br{S}} \|\br{D}\|_* + \lambda \|\br{S}\|_1 + \frac{1}{2}\mu(\|\br{D}\|_F^2 + \|\br{S}\|_F^2)\\
	&s.t. \br{C} = \br{D} + \br{S}, \mu>0,
	\end{aligned} 	
	\end{equation}
	where $\|\cdot \|_*$ is the nuclear norm of a matrix, $\|\cdot \|_1$ is the $l_1$ norm, and $\|\cdot \|_F$ is the Frobenius norm.
	
	In this form, we can solve Eq.~\ref{E11} by using robust PCA~\cite{candes2011robust}. After processing all clusters, the polarization guided model can deliver a pleasing result, which is the intermediate result during the iteration optimization of the specular reflection separation. Since the specular reflection is partially polarized, there are still parts of the specular component remain in the scene, as shown in the first iteration in Fig.~\ref{iteration}. To remove the constant specular component, we reformulate the specular reflection separation to a global energy function with the polarization guided model as an implicit function $f(\cdot)$. The details of the specular reflection separation and the corresponding optimization solution are presented in the next section.

	\section{Specular Reflection Separation}\label{Approach}

	\subsection{Problem Formulation}
	
	According to the dichromatic reflection model~\cite{klinker1988measurement}, the image intensity is a linear combination of diffuse and specular component. However, due to the geometry of the object and the inconstant illumination, the estimation of the corresponding coefficients is an ill-posed problem. In this section, we present a novel method to separate the specular reflection on the basis of both chromaticity and polarization. 
	
	Instead of estimating the coefficients of the specular component and the value of illumination, we directly consider the specular reflection separation as a one-to-two image domain translation problem which aims to extract two components from one given image. This task can be regarded as the following separation model:
	\begin{align}\label{sm}
	\br{I} = \br{R_D} + \br{R_S},
	\end{align}
	where $\br{I}$ is the observation of the input image with specular reflection. $\br{R_D}$ and $\br{R_S}$ are expectant diffuse component and specular component. 
	
	Specular reflection separation aims to reconstruct the $\br{R_D}$ and $\br{R_S}$ from input $\br{I}$. We transform Eq.~\ref{sm} to minimize the following optimization problem with diffuse and specular constraints:
	\begin{equation}\label{E8}
	\begin{aligned}
	&\mathop{min}\limits_{\br{R_D},\br{R_S}} \Phi(\br{R_D})+ \Psi(\br{R_S})\\
	&s.t. \br{I} = \br{R_D} + \br{R_S}, \br{R_S} > 0,
	\end{aligned} 	
	\end{equation}
	where $\br{I}, \br{R_D}$, and $\br{R_S}$ are all matrices with a size of $N\times 3$. $N$ is the total number of pixels and each row represents a pixel color. The subjective term aims to maintain the accuracy of reconstructed components. The two items designate the implicit priors imposed on $\br{R_D}$ and $\br{R_S}$ to regularize inference. Whether these priors can portray the characteristics will play a key role to achieve promising results.
	
	For $\br{R_D}$, the implicit prior $\Phi(\cdot)$ aims to constrain the expectant diffuse component by the polarization guided model $f(\cdot)$. By taking advantage of the polarization observation, we use the result of polarization guided model $f(\br{D})$ to constrain the desired result of diffuse component $\br{R_D}$. We present the detail of the polarization guided model in Sec~\ref{PGM}. During the specular reflection separation, the reconstructed diffuse image $\br{R_D}$ should always be constrained by the polarization-based reconstructed diffuse component $f(\br{D})$. For $\br{R_S}$, the specular component tends to be sparse and non-negative. In this case, we constrain $\br{R_S}$ by applying $l_1$ norm of a matrix, which counts the number of non-zero entries in the matrix. In this case, Eq.11 can be transformed into the following global energy function:
	\begin{align}\label{E2}
	\mathop{min}\limits_{\br{R_D},\br{R_S}} \|\br{I} - \br{R_D} - \br{R_S}\|_F^2 + \|\br{R_D} - f(\br{D})\|_F^2+ \|\br{R_S}\|_1.
	\end{align}
	
	The proposed optimization function needs to decouple the fidelity term and regularization terms. In this work, we use the ADMM strategy~\cite{liu2013linearized} to optimize Eq.~\ref{E2}. In addition, we clamp the negative entries to zero directly during each iteration. To make the optimization tractable, we relax Eq.~\ref{E2} by introducing the augmented Lagrange function for the above optimization problem. The optimization function can further be defined as:
	\begin{align} \label{E3}
	\begin{split}
	&\mathop{min}\limits_{\br{R_D},\br{R_S},\br{D}} 
	\br{S}^T(\br{I} - \br{R_D} - \br{R_S}) +\frac{\rho}{2} \|\br{I} - \br{R_D} - \br{R_S}\|_F^2+\\
	&(\br{S}_{pol}^T(\br{R_D}-f(\br{D})) + \frac{\rho_{pol}}{2}\|\br{R_D}-f(\br{D})\|_F^2)+ \lambda\|\br{R_S}\|_1,
	\end{split}
	\end{align}
	where $\br{S}_{pol}$ and $\rho_{pol}$ are multiplier and penalty parameter for polarization guided model. Correspondingly, $\br{S}$ and $\rho$ are similar parameters for global energy function. $\lambda$ is the parameter used to balance the sparsity of specular component. $f(\br{D})$ is the result of polarization guided model imposed on the desire result of $\br{R_D}$. The $l_1$ norm counts the number of non-zero entries in the specular matrix $\br{R_S}$ and leads to sparse. 
	
	\subsection{Optimization}\label{opt}
	As we mentioned above, with the help of the polarization chromaticity image, the polarization guided model is the implicit function on the desired result of $\br{R_D}$. Based on the ADMM strategy, we can get two sub-problem by splitting the variables from Eq.~\ref{E3}. The sub-problems about $\br{R_D}$ and $\br{R_S}$ can be formulated as:
	\begin{equation} \label{E12}
	\begin{cases}
	\begin{aligned}
	\br{R_D}^{k+1} = &\arg\mathop{min}\limits_{\br{R_D}}\frac{\rho_{pol}}{2}\|\br{R_D} - (f(\br{D})^{k+1} - y_{pol}^{k})\|_F^2 + \\
	&\frac{\rho}{2}\|\br{R_D} - (\br{I} - \br{R_S}^{k} + y^{k}) \|_F^2,\\
	\br{R_S}^{k+1} = &\arg\mathop{min}\limits_{\br{R_S}}\lambda\|\br{R_S}\|_1 + \\
	&\frac{\rho}{2}\|\br{R_S} - (\br{I} - \br{R_D}^{k} + y^{k}) \|_F^2,
	\end{aligned}
	\end{cases}
	\end{equation}
	where $y_{pol}^{k} = (1/\rho_{pol})\br{S}_{pol}^{k}$, $y^{k} = (1/\rho)\br{S}^{k}$ are the scaled Lagrange multipliers. 
	
	The polarization guided model $f(\br{D})^{k+1}$ aims to obtain the approximate result of diffuse reflection $\br{R_D}$ during the iteration optimization. The update of $f(\br{D})^{k+1}$ is expressed as:
	\begin{equation} \label{E13}
	f(\br{D})^{k+1} = f(\br{R_D}^{k}).
	\end{equation}
	By fixing $f(\br{D})^{k+1}$, we can use L-BFGS algorithm~\cite{zhu1997algorithm} to minimize the $\br{R_D}$ in Eq.~\ref{E12} rather than directly calculate derivative of the function with respect to $\br{R_D}$. The optimization can yield better results than the closed-form solution.
	
	After updating $f(\br{D})^{k+1}$ and $\br{R_D}^{k+1}$, $\br{R_S}$ in Eq.~\ref{E12} can be calculated by the proximal operator of the $l1$ norm~\cite{parikh2014proximal}:
	\begin{equation} \label{E14}
	\begin{aligned}
	&\br{R_S}^{k+1} = S_{\lambda/\rho}(\br{I} - \br{R_D}^{k} + y^{k})\\
	&\lambda = 1/\sqrt{N_s},
	\end{aligned}
	\end{equation}
	where $S$ is the proximal operator, $N_s$ is the number of pixels in $\br{R_S}$. We set $\lambda$ to be dynamically changed with the number of pixels among the specular reflection during the iteration. 
	
	The penalty parameters $\rho_{pol}$ and $\rho$ are initialized to 1.1 and updated at a multiple of 1.05. $\br{S}_{pol}$ and $\br{S}$ are updated as follows:
	\begin{equation} \label{E15}
	\begin{cases}
	{\br{S}_{pol}^{k+1}} = \br{S}_{pol}^{k} + \rho_{pol}(\br{R_D}^{k} - \br{D}^{k}),\\
	{\br{S}^{k+1}} = \br{S}^{k} + \rho(\br{I} - \br{R_D}^{k} - \br{R_S}^{k}).
	\end{cases}
	\end{equation}
	
	\begin{algorithm}[h] 
		\caption{Polarization Guided Specular Reflection Separation} 
		\label{Alg} 
		\begin{algorithmic}[1] 
			\Require The observation $\br{I}$, $\rho_{pol}$, $\rho$, maxiter = 50, $\epsilon = 10^{-3}$;
			\State Calculate the polarization chromaticity image $I_{chro}$ by Eq.~\ref{E7};
			\State initialization: $\br{R_D} = I_{rawD}$, $\br{R_S} = I_{rawS}$ according to Eq.~\ref{E6} ; 
			\Repeat 
			\State Updating $f(\br{D})$ by Eq.~\ref{E13}; 
			\State Updating $\br{R_D}$ by Eq.~\ref{E12}; 
			\State Updating $\br{R_S}$ by Eq.~\ref{E14}; 
			\State Updating $\br{S}_{pol}$ and $\br{S}$ by Eq.~\ref{E15}; 
			\State Break: $\left\{ \|\br{S}_{pol}^{k+1} - \br{S}_{pol}^{k}\| < \epsilon \hspace{1.5mm} \& \hspace{1.5mm} \|\br{S}^{k+1} - \br{S}^{k}\|  < \epsilon \right\}$;
			\Until
			\Ensure $\br{R_D}\in\mathbb{R}^{N\times3}$  , $\br{R_S}\in\mathbb{R}^{N\times3}$
		\end{algorithmic} 
	\end{algorithm}
	
	Overall, the optimization of the global energy function is summarized in Algorithm.~\ref{Alg}. In order to visually demonstrate the effectiveness and performance of iteration optimization, we show the first and final iterations of the optimization. As shown in Fig.~\ref{iteration}, the proposed method effectively separate the specular reflection during the iteration of optimization, regardless of the effect of illumination. More experimental results are shown in Sec.~\ref{Ex}.
	
	\section{Experiments and Discussions}\label{Ex}
	In this section, we collect a dataset with different illumination to evaluate the effectiveness and robustness of the proposed method. We then compare our method with some outstanding methods by visual evaluation and quantitative evaluation. At last, we do a robustness analysis of our method under different illumination.
	\begin{figure*}[htbp]	
		\centering
		\includegraphics[width=1\linewidth]{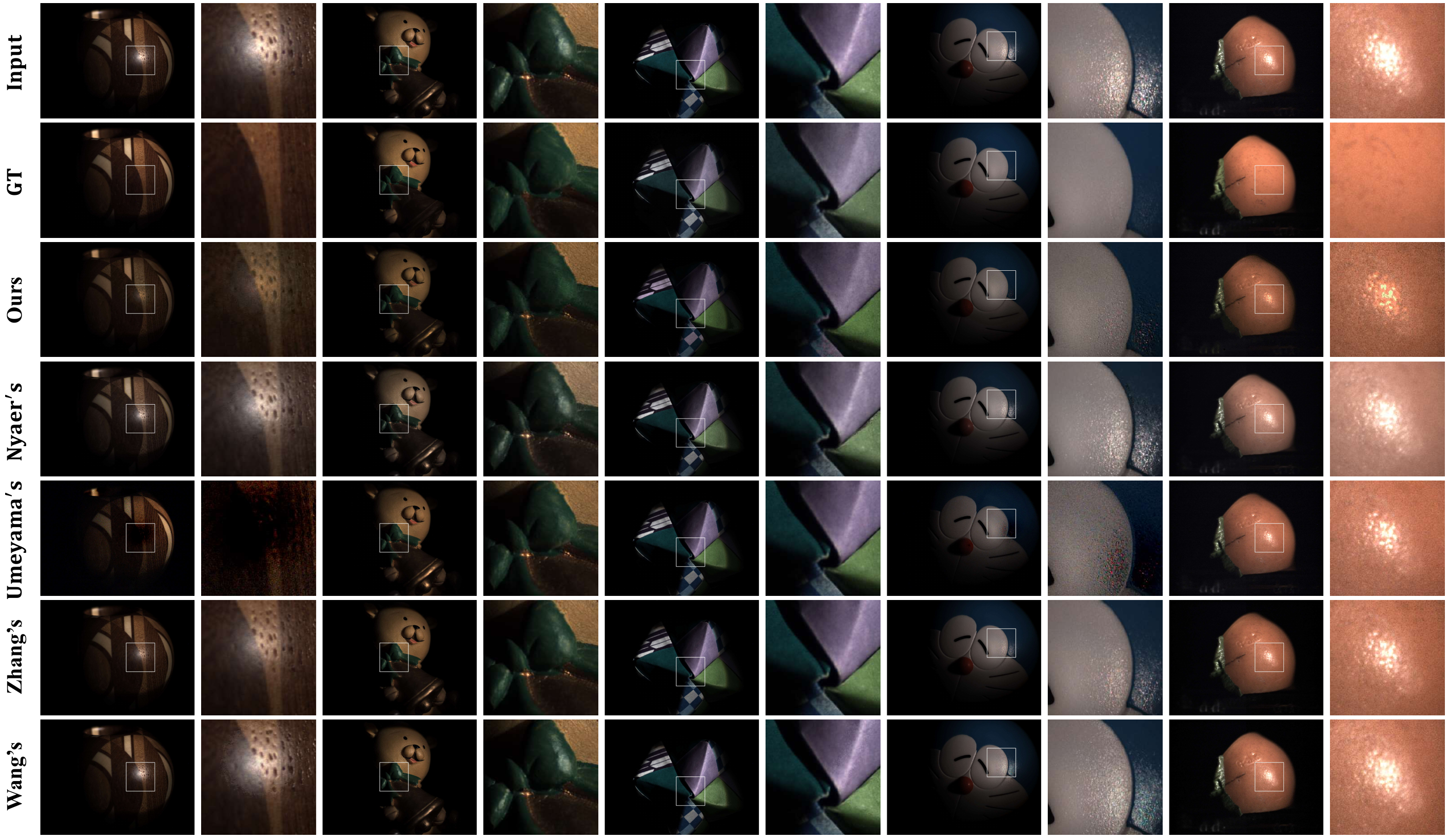}
		\caption{We compare our method to those polarization-based methods of Nayar~\etal~\cite{nayar1997separation}, Umeyama~\etal~\cite{umeyama2004separation}, Zhang~\etal~\cite{zhang2011reflection}, and Wang~\etal~\cite{wang2017specularity}.}
		\label{compare_pol}
	\end{figure*}

	\subsection{Dataset}\label{es}
	The ideal scene with pure-diffuse reflection does not exist in the real world. Former methods typically added simulated specular reflection to obtain the dataset for comparison. Instead of simulating the pseudo data, we intend to capture the ground truth for the comparisons. To achieve this purpose, we add a polarizer in front of the point light source and rotate the polarizer until there is no specular reflection on the visualization. However, due to the diverging rays of light, rotating the polarizer cannot thoroughly eliminate the specular reflection. Moreover, adding a polarizer in front of the light will slightly change the color of the light source. Despite all these, the captured ground truth can still be considered as the reference to effectively and intuitively help us with the quantitative comparison. According to the newly released single-chip polarized color sensor and corresponding demosaicing method~\cite{wen2019convolutional}, it is able to simultaneously capture RGB and polarimetric information of the scene. By using such a sensor, we collect a dataset that includes 8 scenes. Each of them is captured under 7 different illumination for experimental evaluation. The size of the captured images is $1032\times1384$. The dataset will be released soon.
	
	\subsection{Comparison Experiments}
	
	In order to have clear visual comparison experiments, we classify the specular removal methods into two groups: the one with polarization and the other with chromaticity. In addition, we also compare our method with all of them by quantitative analysis. 
	
	\subsubsection{Specular removal with polarization}
	
	The proposed method is not the first to introduce polarization into the specular highlight removal. Zhang \etal~\cite{zhang2011reflection} directly utilize polarization on each of R, G, B channels by using TRS. As we mentioned in \textit{Analysis of the polarization}, the specular reflection is partially polarized. Therefore, the constant part of the specular component cannot be separated by TRS. Nayar \etal~\cite{nayar1997separation} use color and polarimetric information simultaneously to constrain each pixel in RGB space. Apart from its heavy computational burden, the assumption and threshold in their work limit the performance. Umeyama~\etal~\cite{umeyama2004separation} install polarization into the specular removal algorithm with the fixed coefficient. In this case, the specular component cannot be fully removed. Wang~\etal~\cite{wang2017specularity} improve Umeyama's method by replacing the fixed coefficients with the spatially variable mixing coefficients. These methods cannot handle many scenes well in the real world and consume lots of time. The comparison results show that our method performs better, as shown in Fig.~\ref{compare_pol}.
	\begin{figure*}[htbp]	
		\centering
		\includegraphics[width=1\linewidth]{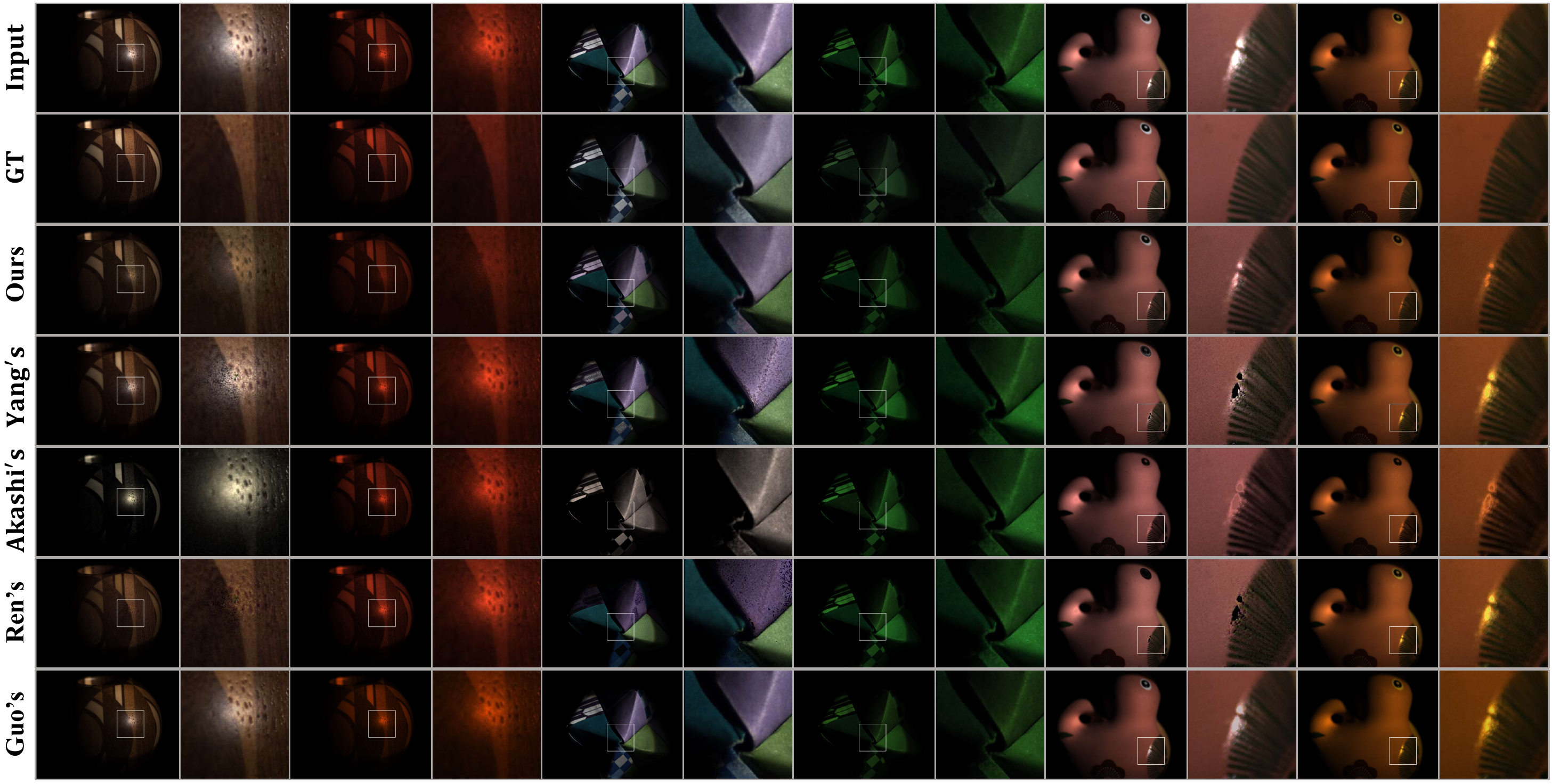}
		\caption{We compare our method to those chromaticity-based methods of Yang~\etal~\cite{yang2010real}, Akashi~\etal~\cite{akashi2014separation}, Ren~\etal~\cite{ren2017specular}, and Guo~\etal~\cite{guo2018single}.}
		\label{compare_single}
	\end{figure*}
	\begin{table}[h]
		\fontsize{9}{12}\selectfont
		\renewcommand\tabcolsep{10pt} 
		\begin{center}
			\caption{Quantitative comparison of separation results. The best and the second results are in \textcolor{red}{red} and \textcolor{blue}{blue} fonts.}
			\label{QA}
			\begin{tabular}{l|cccc}
				\toprule\hline
				Method & PSNR & SSIM & CA &SD \\
				\hline
				Nayar~\cite{nayar1997separation} &30.131 &\textcolor{red}{0.887} &22.100 &\textcolor{blue}{0.037}  \\
				Umeyama~\cite{umeyama2004separation} &26.142 &0.649 &19.655 &0.042  \\
				Yang~\cite{yang2010real} &31.585 &0.845 &23.834 &0.037  \\
				Zhang~\cite{zhang2011reflection} &31.775 &0.814 &\textcolor{red}{25.004} &0.039 \\
				Akashi~\cite{akashi2014separation} &30.157 &0.737 &24.098 &0.061 \\
				Wang~\cite{wang2017specularity} &31.454 &0.726 &24.429 &0.039 \\
				Ren~\cite{ren2017specular} &\textcolor{blue}{31.764} &0.858 &24.548 &\textcolor{blue}{0.037}  \\
				Guo~\cite{guo2018single} &30.176  &0.854 &23.869 &0.040 \\
				Ours &\textcolor{red}{32.097} &\textcolor{blue}{0.877} &\textcolor{blue}{24.909} &\textcolor{red}{0.034} \\	
				\hline\bottomrule
			\end{tabular}
		\end{center}
	\end{table}
	\begin{figure*}[htbp]	
		\centering
		\includegraphics[width=1\linewidth]{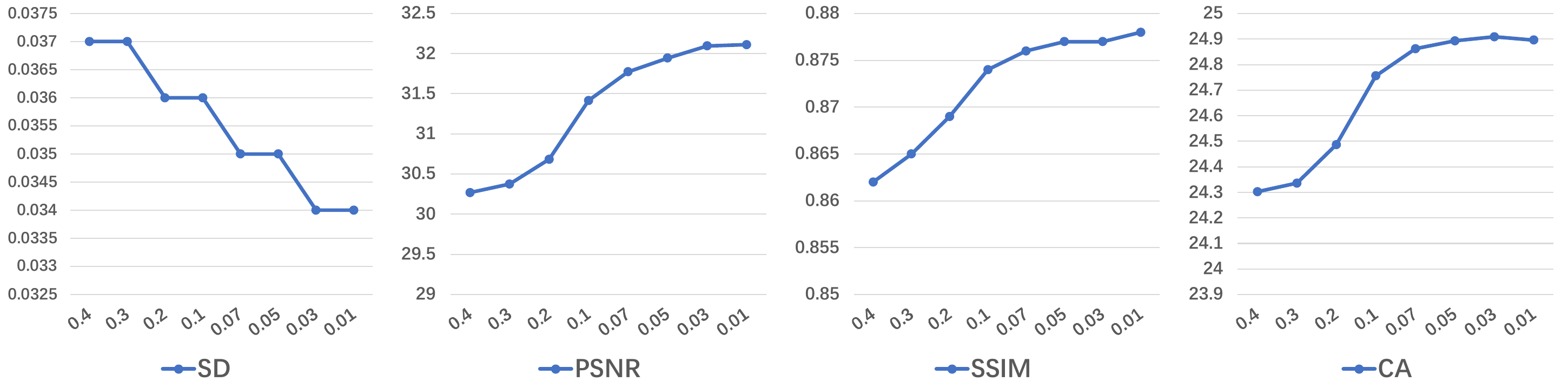}
		\caption{The value of evaluation metrics with different chromatic threshold selections.}
		\label{T}
	\end{figure*}

	\subsubsection{Specular removal with chromaticity}
	
	We compare our results with those of methods based on the chromaticity. Yang~\etal~\cite{yang2010real} and Akashi~\etal~\cite{akashi2014separation} formulate the specular removal into a energy function. Due to the assumption of illumination chromaticity, their methods cannot stably separate the specular reflection. Ren~\etal~\cite{ren2017specular} propose an approach that relies on the color-line. It can achieve better results under white illumination, as shown in Fig.~\ref{compare_single}. Yet the performance of their method will decrease with the complexity of the scene, such as the scenes under different colors of illumination. Guo~\etal~\cite{guo2018single} propose a sparse and low-rank reflection model without the knowledge of illumination. However, due to the partial polarized specular component, their method cannot fully remove the specular component. Although all these methods successfully remove most of the highlights in the ideal condition, ambient lighting usually changes and is difficult to estimate. In this regard, our method can still achieve good performance as shown in Fig.~\ref{compare_single}. 
	
	\subsubsection{Quantitative comparison}
	
	\begin{figure*}[htbp]	
		\centering
		\includegraphics[width=1\linewidth]{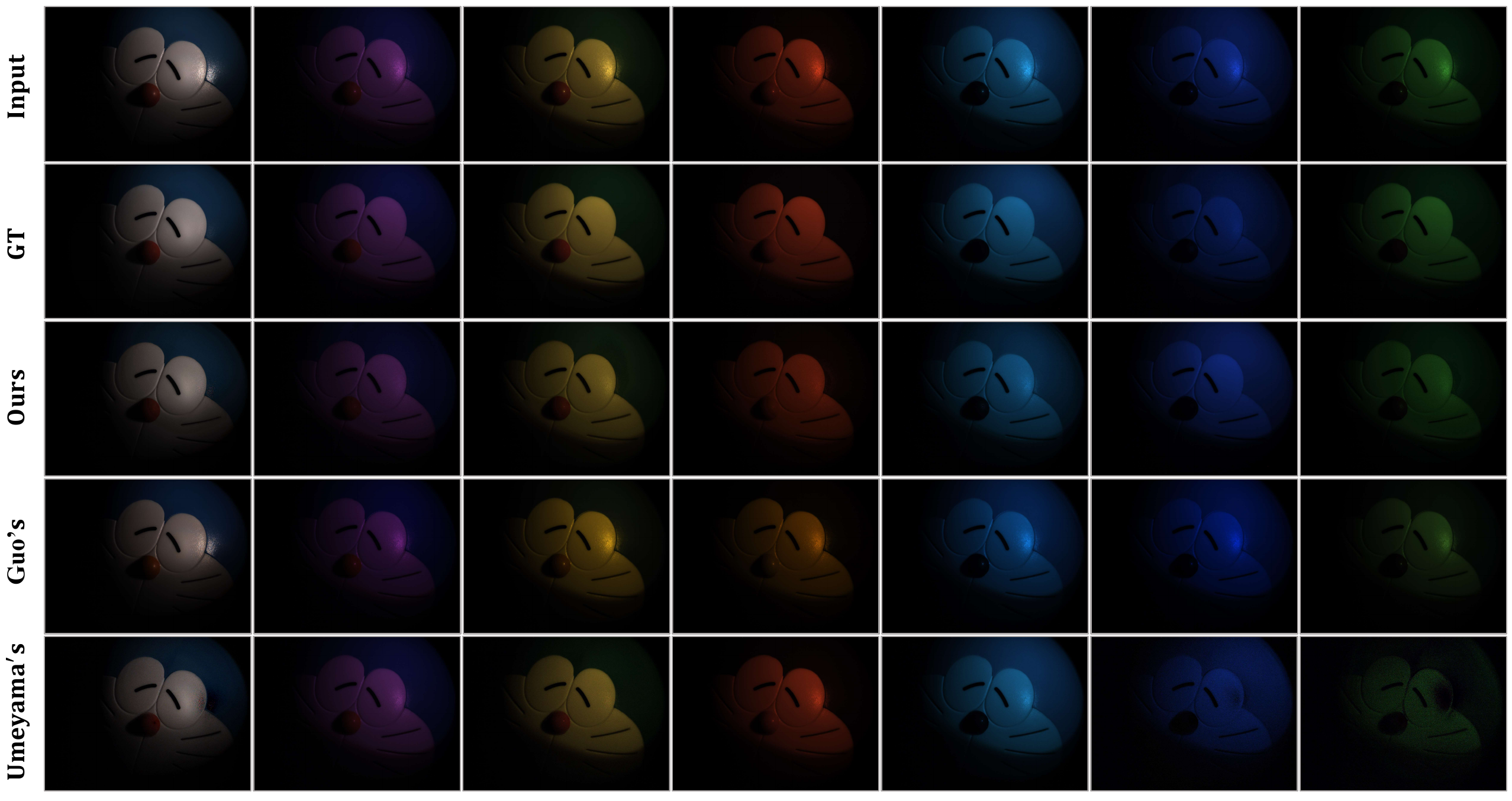}
		\caption{The effect of illumination chromaticity on specular reflection separation. We compare our method to those of Guo~\etal~\cite{guo2018single} and Umeyama~\etal~\cite{umeyama2004separation}.}
		\label{illumination}
	\end{figure*}
	\begin{figure*}[htbp]	
		\centering
		\includegraphics[width=1\linewidth]{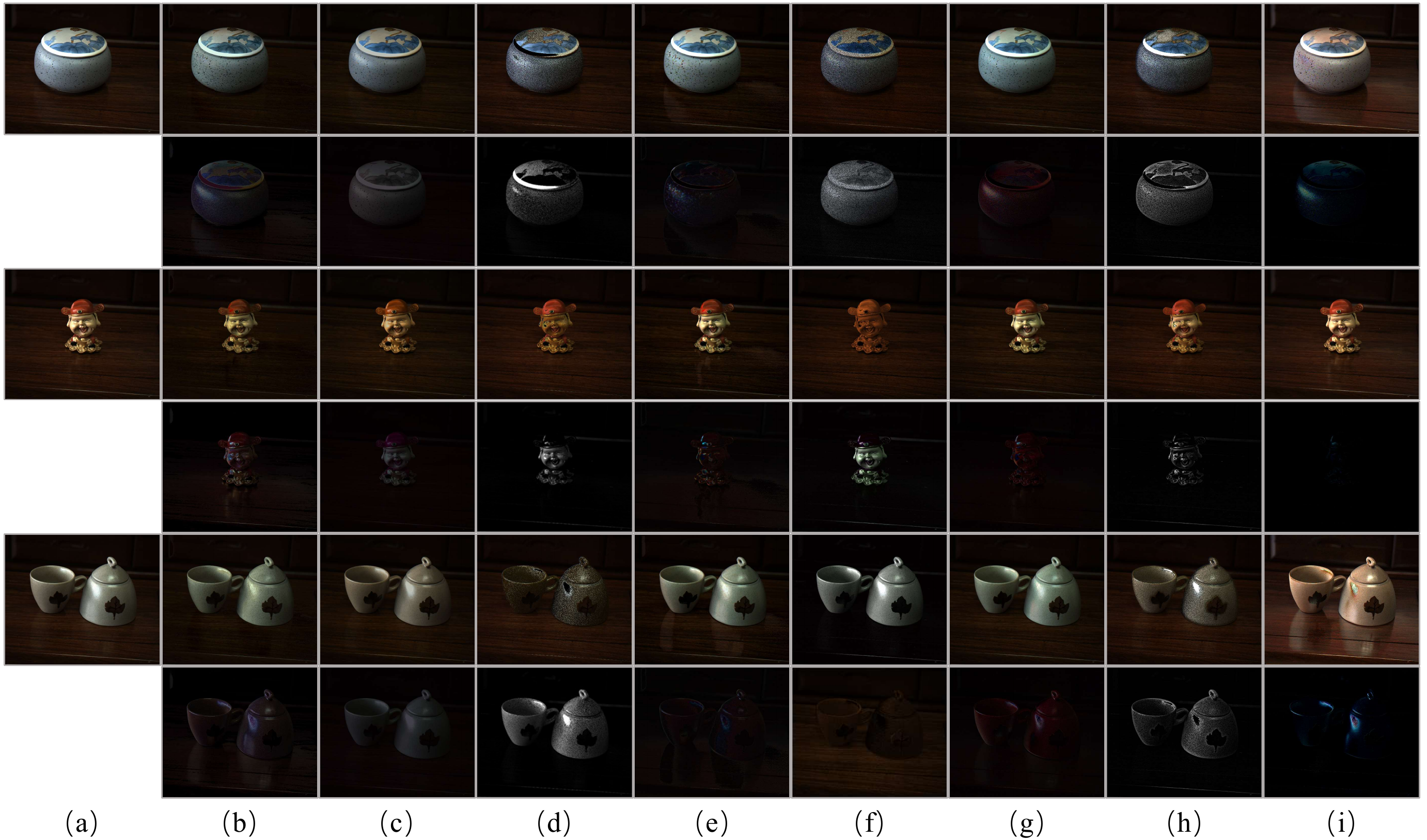}
		\caption{Visual comparison for three real images taken under natural illumination. (a) Input image, (b) separation results of ours, (c) separation results of Guo~\etal~\cite{guo2018single}, (d) separation results of Ren~\etal~\cite{ren2017specular}, (e) separation results of Wang~\etal~\cite{wang2017specularity}, (f) separation results of Akashi~\etal~\cite{akashi2014separation}, (g) separation results of Zhang~\etal~\cite{zhang2011reflection}, (h) separation results of Yang~\etal~\cite{yang2010real}, (i) separation results of Umeyama~\etal~\cite{umeyama2004separation}.}
		\label{real}
	\end{figure*}
	
	The SSIM~\cite{wang2004image} and PSNR are commonly used as the metrics for the quantitative comparison of different methods. However, the ideal ground truth is not accessible in real scenarios. For the ground truth we captured, these metrics are not enough to evaluate the performance of the specular separation methods. To achieve a more precise evaluation, we also introduce Standard Deviation (SD) of the histogram distribution~\cite{wang2017specularity} and Color Accuracy (CA)~\cite{wen2019convolutional} as evaluation metrics. The CA can help to evaluate if there is any color distortion during the specular separation. According to Wang~\cite{wang2017specularity}, the histogram of hue values with weak specular reflection is more concentrated than the one with strong specular reflection. The SD of the hue values of separation result can thus be used to evaluate the performance of specularity removal. Be noted that, unlike other metrics, the better performance of specular separation leads to the smaller SD. The average PSNR, SSIM, CA, and SD are calculated with the captured ground truth of the dataset. As shown in TABLE~\ref{QA}, the proposed method outperforms most methods in all metrics, especially in SD. Though the method of Nayar~\cite{nayar1997separation} has a higher SSIM value and the method of Zhang~\cite{zhang2011reflection} has a higher CA value, but they obtain low value on other metrics because of incomplete separation. Quantitative comparison proves that our method can achieve pleasing separation results while preserving structural information, tiny details, and color information.
	
	\subsubsection{Saturation analysis}
	As shown in the results of comparison experiments, most of the saturated highlight areas cannot be processed well by either the polarization-based methods or chromaticity-based methods. However, the proposed method can still outperform the former methods in the saturation area. Though the proposed method can perform well even in complex scenarios, it cannot thoroughly remove the saturated specular reflection, which will be our further research.
	
	\subsection{Performance Analysis}
	
	\subsubsection{Chromatic threshold selection}
	For the parameter of the polarization guided model, the value of the chromatic threshold should be investigated under different values. The chromatic threshold aims to cluster pixels with similar intrinsic diffuse color. We carried out the experiments on our evaluation metrics to show the effects of different parameters. We tested the performance with $T\in\left\{ 0.4, 0.3, 0.2, 0.1, 0.07, 0.05, 0.03, 0.01 \right\}$. The average evaluation metrics values calculated on the dataset are shown in Fig.~\ref{T}. One can see that better performance can be obtained with a smaller $T$. When the value of $T$ is smaller than 0.03, the improvement is limited while costing a longer running time. In our experiments, we set $T$ to be 0.03 since it achieves better and more stable results.
	
	\subsubsection{Study on illumination}
	To demonstrate the robustness of our method under different colors of illumination, we provide the study on illumination. As we mentioned earlier, most of the existing methods are obliged to estimate or assume illumination. Differently, Guo~\etal~\cite{guo2018single} propose a specular removal method without knowing the illumination. Umeyama~\etal~\cite{umeyama2004separation} use polarization to avoid the requirement of the illumination. To make fair comparisons, we only compare ours with these two methods in the same scene under seven different illumination. When the color of the diffuse component is similar to the illumination (the color of specular component), these two methods will suffer from the color distortion of separation results. As shown in the 3rd, 4th, and 5th rows of Fig.~\ref{illumination}, the results of visual comparisons show that our method can achieve better results under different colors of illumination. 
	
	\subsubsection{Specular separation on real scene}
	As we mentioned in Sec.~\ref{es}, our dataset is captured under a point light source. To demonstrate the performance of our method on the real scene, Fig.~\ref{real} shows the visual comparison for three images taken under natural illumination. The methods in \cite{ren2017specular}, \cite{akashi2014separation}, \cite{umeyama2004separation} and \cite{yang2010real} cannot correctly separate the specular component under natural illumination. Without the calibration of illumination, some diffuse pixels are considered and separated as specular pixels by these methods. The methods in \cite{guo2018single}, \cite{wang2017specularity} and \cite{zhang2011reflection} can remove the specular reflection of these images, however the performance is limited. With the polarization guided model, our method can separate specular reflection well without causing color distortion, even with saturation existed in real scenes.
	
	\section{Conclusions}\label{conclusion}
	
	In this paper, we present an efficient specular reflection separation method with polarization. Through the analysis of the polarization, we develop a polarization guided model to generate the polarization chromaticity image which can accurately cluster the pixels with a similar color of diffuse reflection. The proposed model can separate the specular component for all clusters and obtain an implicit prior to constrain the expectant diffuse reflection image. With the polarization guided model, we further reformulate the problem into a global energy function. By adopting the ADMM strategy, we optimize the energy function to deliver truthful and faithful results of specular reflection separation. It should be noted that our model with polarimetric information will not be influenced by illumination. At last, we conduct extensive experiments to show that the proposed method can work well on the real scene and outperforms the state-of-the-art methods. Our future work is to solve the
	limitations and apply our model to practical applications, including shape from polarization~\cite{rahmann2001reconstruction}, segmentation and detection with glint~\cite{hedley2005simple}, and shape and reflectance analysis~\cite{liu2020separate,cao2019makeup}.

	\ifCLASSOPTIONcaptionsoff
	\newpage
	\fi
	
	
	
	%
	%
	%
	\bibliographystyle{IEEEtran}
	\bibliography{egbib}

\begin{thebibliography}{10}
\providecommand{\url}[1]{#1}
\csname url@samestyle\endcsname
\providecommand{\newblock}{\relax}
\providecommand{\bibinfo}[2]{#2}
\providecommand{\BIBentrySTDinterwordspacing}{\spaceskip=0pt\relax}
\providecommand{\BIBentryALTinterwordstretchfactor}{4}
\providecommand{\BIBentryALTinterwordspacing}{\spaceskip=\fontdimen2\font plus
\BIBentryALTinterwordstretchfactor\fontdimen3\font minus
  \fontdimen4\font\relax}
\providecommand{\BIBforeignlanguage}[2]{{%
\expandafter\ifx\csname l@#1\endcsname\relax
\typeout{** WARNING: IEEEtran.bst: No hyphenation pattern has been}%
\typeout{** loaded for the language `#1'. Using the pattern for}%
\typeout{** the default language instead.}%
\else
\language=\csname l@#1\endcsname
\fi
#2}}
\providecommand{\BIBdecl}{\relax}
\BIBdecl

\bibitem{shafer1985using}
S.~A. Shafer, ``Using color to separate reflection components,'' \emph{Color
  Research \& Application}, vol.~10, no.~4, pp. 210--218, 1985.

\bibitem{lee1990modeling}
H.-C. Lee, E.~J. Breneman, and C.~P. Schulte, ``Modeling light reflection for
  computer color vision,'' \emph{IEEE Transactions on Pattern Analysis and
  Machine Intelligence}, vol.~12, no.~4, pp. 402--409, 1990.

\bibitem{dai2016instance}
J.~Dai, K.~He, and J.~Sun, ``Instance-aware semantic segmentation via
  multi-task network cascades,'' in \emph{Proceedings of the IEEE Conference on
  Computer Vision and Pattern Recognition}, 2016, pp. 3150--3158.

\bibitem{zhu2019depth}
D.~Zhu and W.~A. Smith, ``Depth from a polarisation+ rgb stereo pair,'' in
  \emph{Proceedings of the IEEE Conference on Computer Vision and Pattern
  Recognition}, 2019, pp. 7586--7595.

\bibitem{chen2020learning}
D.~Chen, J.~Li, Z.~Wang, and K.~Xu, ``Learning canonical shape space for
  category-level 6d object pose and size estimation,'' in \emph{Proceedings of
  the IEEE/CVF Conference on Computer Vision and Pattern Recognition}, 2020,
  pp. 11\,973--11\,982.

\bibitem{cui2017polarimetric}
Z.~Cui, J.~Gu, B.~Shi, P.~Tan, and J.~Kautz, ``Polarimetric multi-view
  stereo,'' in \emph{Proceedings of the IEEE Conference on Computer Vision and
  Pattern Recognition}, 2017, pp. 1558--1567.

\bibitem{singla2014motion}
N.~Singla, ``Motion detection based on frame difference method,''
  \emph{International Journal of Information \& Computation Technology},
  vol.~4, no.~15, pp. 1559--1565, 2014.

\bibitem{tominaga1991surface}
S.~Tominaga, ``Surface identification using the dichromatic reflection model,''
  \emph{IEEE Transactions on Pattern Analysis \& Machine Intelligence}, no.~7,
  pp. 658--670, 1991.

\bibitem{lu2015intensity}
F.~Lu, Y.~Matsushita, I.~Sato, T.~Okabe, and Y.~Sato, ``From intensity profile
  to surface normal: photometric stereo for unknown light sources and isotropic
  reflectances,'' \emph{IEEE transactions on pattern analysis and machine
  intelligence}, vol.~37, no.~10, pp. 1999--2012, 2015.

\bibitem{lu2015uncalibrated}
F.~Lu, I.~Sato, and Y.~Sato, ``Uncalibrated photometric stereo based on
  elevation angle recovery from brdf symmetry of isotropic materials,'' in
  \emph{Proceedings of the IEEE Conference on Computer Vision and Pattern
  Recognition}, 2015, pp. 168--176.

\bibitem{klinker1990physical}
G.~J. Klinker, S.~A. Shafer, and T.~Kanade, ``A physical approach to color
  image understanding,'' \emph{International Journal of Computer Vision},
  vol.~4, no.~1, pp. 7--38, 1990.

\bibitem{yang2012new}
J.~Yang, Z.~Cai, L.~Wen, Z.~Lei, G.~Guo, and S.~Z. Li, ``A new projection space
  for separation of specular-diffuse reflection components in color images,''
  in \emph{Asian Conference on Computer Vision}.\hskip 1em plus 0.5em minus
  0.4em\relax Springer, 2012, pp. 418--429.

\bibitem{yang2013separating}
J.~Yang, L.~Liu, and S.~Li, ``Separating specular and diffuse reflection
  components in the hsi color space,'' in \emph{Proceedings of the IEEE
  International Conference on Computer Vision Workshops}, 2013, pp. 891--898.

\bibitem{yang2014efficient}
Q.~Yang, J.~Tang, and N.~Ahuja, ``Efficient and robust specular highlight
  removal,'' \emph{IEEE transactions on pattern analysis and machine
  intelligence}, vol.~37, no.~6, pp. 1304--1311, 2014.

\bibitem{akashi2014separation}
Y.~Akashi and T.~Okatani, ``Separation of reflection components by sparse
  non-negative matrix factorization,'' in \emph{Asian Conference on Computer
  Vision}.\hskip 1em plus 0.5em minus 0.4em\relax Springer, 2014, pp. 611--625.

\bibitem{bajcsy1996detection}
R.~Bajcsy, S.~W. Lee, and A.~Leonardis, ``Detection of diffuse and specular
  interface reflections and inter-reflections by color image segmentation,''
  \emph{International Journal of Computer Vision}, vol.~17, no.~3, pp.
  241--272, 1996.

\bibitem{shen2008chromaticity}
H.-L. Shen, H.-G. Zhang, S.-J. Shao, and J.~H. Xin, ``Chromaticity-based
  separation of reflection components in a single image,'' \emph{Pattern
  Recognition}, vol.~41, no.~8, pp. 2461--2469, 2008.

\bibitem{article}
R.~Tan and K.~Ikeuchi, ``Separating reflection components of textured surfaces
  using a single image,'' \emph{IEEE transactions on pattern analysis and
  machine intelligence}, vol.~27, pp. 178--93, 03 2005.

\bibitem{yang2010real}
Q.~Yang, S.~Wang, and N.~Ahuja, ``Real-time specular highlight removal using
  bilateral filtering,'' in \emph{European conference on computer
  vision}.\hskip 1em plus 0.5em minus 0.4em\relax Springer, 2010, pp. 87--100.

\bibitem{huard1997polarization}
S.~Huard, ``Polarization of light,'' \emph{Polarization of Light, by Serge
  Huard, pp. 348. ISBN 0-471-96536-7. Wiley-VCH, January 1997.}, p. 348, 1997.

\bibitem{wolff1993constraining}
L.~B. Wolff and T.~E. Boult, ``Constraining object features using a
  polarization reflectance model,'' \emph{Physics-Based Vision: Principles and
  Practice: Radiometry}, vol.~1, p. 167, 1993.

\bibitem{born2013principles}
M.~Born and E.~Wolf, \emph{Principles of optics: electromagnetic theory of
  propagation, interference and diffraction of light}.\hskip 1em plus 0.5em
  minus 0.4em\relax Elsevier, 2013.

\bibitem{nayar1997separation}
S.~K. Nayar, X.-S. Fang, and T.~Boult, ``Separation of reflection components
  using color and polarization,'' \emph{International Journal of Computer
  Vision}, vol.~21, no.~3, pp. 163--186, 1997.

\bibitem{umeyama2004separation}
S.~Umeyama and G.~Godin, ``Separation of diffuse and specular components of
  surface reflection by use of polarization and statistical analysis of
  images,'' \emph{IEEE Transactions on Pattern Analysis and Machine
  Intelligence}, vol.~26, no.~5, pp. 639--647, 2004.

\bibitem{wang2017specularity}
F.~Wang, S.~Ainouz, C.~Petitjean, and A.~Bensrhair, ``Specularity removal: a
  global energy minimization approach based on polarization imaging,''
  \emph{Computer Vision and Image Understanding}, vol. 158, pp. 31--39, 2017.

\bibitem{guo2018single}
J.~Guo, Z.~Zhou, and L.~Wang, ``Single image highlight removal with a sparse
  and low-rank reflection model,'' in \emph{Proceedings of the European
  Conference on Computer Vision (ECCV)}, 2018, pp. 268--283.

\bibitem{candes2011robust}
E.~J. Cand{\`e}s, X.~Li, Y.~Ma, and J.~Wright, ``Robust principal component
  analysis?'' \emph{Journal of the ACM (JACM)}, vol.~58, no.~3, p.~11, 2011.

\bibitem{khan2017analytical}
H.~A. Khan, J.-B. Thomas, and J.~Y. Hardeberg, ``Analytical survey of highlight
  detection in color and spectral images,'' in \emph{International Workshop on
  Computational Color Imaging}.\hskip 1em plus 0.5em minus 0.4em\relax
  Springer, 2017, pp. 197--208.

\bibitem{artusi2011survey}
A.~Artusi, F.~Banterle, and D.~Chetverikov, ``A survey of specularity removal
  methods,'' in \emph{Computer Graphics Forum}, vol.~30, no.~8.\hskip 1em plus
  0.5em minus 0.4em\relax Wiley Online Library, 2011, pp. 2208--2230.

\bibitem{klinker1987using}
G.~J. Klinker, S.~A. Shafer, and T.~Kanade, ``Using a color reflection model to
  separate highlights from object color,'' in \emph{Proc. ICCV}, vol.~87.\hskip
  1em plus 0.5em minus 0.4em\relax Citeseer, 1987, pp. 145--150.

\bibitem{klinker1988measurement}
G.~Klinker, S.~Shafer, and T.~Kanade, ``The measurement of highlights in color
  images,'' vol.~2, 07 2004.

\bibitem{schluns1995analysis}
K.~Schl{\"u}ns and M.~Teschner, ``Analysis of 2d color spaces for highlight
  elimination in 3d shape reconstruction,'' in \emph{Proc. ACCV}, vol.~2, 1995,
  pp. 801--805.

\bibitem{schluns1995fast}
K.~Schlns and M.~Teschner, ``Fast separation of reflection components and its
  application in 3d shape recovery,'' 02 1997.

\bibitem{tan2006separation}
P.~Tan, L.~Quan, and S.~Lin, ``Separation of highlight reflections on textured
  surfaces,'' in \emph{2006 IEEE Computer Society Conference on Computer Vision
  and Pattern Recognition (CVPR'06)}, vol.~2.\hskip 1em plus 0.5em minus
  0.4em\relax IEEE, 2006, pp. 1855--1860.

\bibitem{yoon2006fast}
K.-J. Yoon, Y.~Choi, and I.~S. Kweon, ``Fast separation of reflection
  components using a specularity-invariant image representation,'' in
  \emph{2006 International Conference on Image Processing}.\hskip 1em plus
  0.5em minus 0.4em\relax IEEE, 2006, pp. 973--976.

\bibitem{shen2009simple}
H.-L. Shen and Q.-Y. Cai, ``Simple and efficient method for specularity removal
  in an image,'' \emph{Applied optics}, vol.~48, no.~14, pp. 2711--2719, 2009.

\bibitem{mallick2006specularity}
S.~P. Mallick, T.~Zickler, P.~N. Belhumeur, and D.~J. Kriegman, ``Specularity
  removal in images and videos: A pde approach,'' in \emph{European Conference
  on Computer Vision}.\hskip 1em plus 0.5em minus 0.4em\relax Springer, 2006,
  pp. 550--563.

\bibitem{tappen2003recovering}
M.~F. Tappen, W.~T. Freeman, and E.~H. Adelson, ``Recovering intrinsic images
  from a single image,'' in \emph{Advances in neural information processing
  systems}, 2003, pp. 1367--1374.

\bibitem{suo2016fast}
J.~Suo, D.~An, X.~Ji, H.~Wang, and Q.~Dai, ``Fast and high quality highlight
  removal from a single image,'' \emph{IEEE Transactions on Image Processing},
  vol.~25, no.~11, pp. 5441--5454, 2016.

\bibitem{liu2015saturation}
Y.~Liu, Z.~Yuan, N.~Zheng, and Y.~Wu, ``Saturation-preserving specular
  reflection separation,'' in \emph{Proceedings of the IEEE Conference on
  Computer Vision and Pattern Recognition}, 2015, pp. 3725--3733.

\bibitem{ren2017specular}
W.~Ren, J.~Tian, and Y.~Tang, ``Specular reflection separation with color-lines
  constraint,'' \emph{IEEE Transactions on Image Processing}, vol.~26, no.~5,
  pp. 2327--2337, 2017.

\bibitem{lee1992detection}
S.~W. Lee and R.~Bajcsy, ``Detection of specularity using color and multiple
  views,'' in \emph{European Conference on Computer Vision}.\hskip 1em plus
  0.5em minus 0.4em\relax Springer, 1992, pp. 99--114.

\bibitem{sato1994temporal}
Y.~Sato and K.~Ikeuchi, ``Temporal-color space analysis of reflection,''
  \emph{JOSA A}, vol.~11, no.~11, pp. 2990--3002, 1994.

\bibitem{lin2001separation}
S.~Lin and H.-Y. Shum, ``Separation of diffuse and specular reflection in color
  images,'' in \emph{Proceedings of the 2001 IEEE Computer Society Conference
  on Computer Vision and Pattern Recognition. CVPR 2001}, vol.~1.\hskip 1em
  plus 0.5em minus 0.4em\relax IEEE, 2001, pp. I--I.

\bibitem{chen2006mesostructure}
T.~Chen, M.~Goesele, and H.-P. Seidel, ``Mesostructure from specularity,'' in
  \emph{2006 IEEE Computer Society Conference on Computer Vision and Pattern
  Recognition (CVPR'06)}, vol.~2.\hskip 1em plus 0.5em minus 0.4em\relax IEEE,
  2006, pp. 1825--1832.

\bibitem{lin2002diffuse}
S.~Lin, Y.~Li, S.~B. Kang, X.~Tong, and H.-Y. Shum, ``Diffuse-specular
  separation and depth recovery from image sequences,'' in \emph{European
  conference on computer vision}.\hskip 1em plus 0.5em minus 0.4em\relax
  Springer, 2002, pp. 210--224.

\bibitem{yang2010uniform}
Q.~Yang, S.~Wang, N.~Ahuja, and R.~Yang, ``A uniform framework for estimating
  illumination chromaticity, correspondence, and specular reflection,''
  \emph{IEEE Transactions on Image Processing}, vol.~20, no.~1, pp. 53--63,
  2010.

\bibitem{weiss2001deriving}
Y.~Weiss, ``Deriving intrinsic images from image sequences,'' in
  \emph{Proceedings Eighth IEEE International Conference on Computer Vision.
  ICCV 2001}, vol.~2.\hskip 1em plus 0.5em minus 0.4em\relax IEEE, 2001, pp.
  68--75.

\bibitem{agrawal2005removing}
A.~Agrawal, R.~Raskar, S.~K. Nayar, and Y.~Li, ``Removing photography artifacts
  using gradient projection and flash-exposure sampling,'' in \emph{ACM
  SIGGRAPH 2005 Papers}, 2005, pp. 828--835.

\bibitem{hyvarinen2000independent}
A.~Hyv{\"a}rinen and E.~Oja, ``Independent component analysis: algorithms and
  applications,'' \emph{Neural networks}, vol.~13, no. 4-5, pp. 411--430, 2000.

\bibitem{zhang2011reflection}
L.~Zhang, E.~R. Hancock, and G.~A. Atkinson, ``Reflection component separation
  using statistical analysis and polarisation,'' in \emph{Iberian Conference on
  Pattern Recognition and Image Analysis}.\hskip 1em plus 0.5em minus
  0.4em\relax Springer, 2011, pp. 476--483.

\bibitem{brewster1815laws}
D.~Brewster, ``On the laws which regulate the polarisation of light by
  reflexion from transparent bodies,'' \emph{Philosophical Transactions of the
  Royal Society of London}, vol. 105, pp. 125--159, 1815.

\bibitem{winthrop1965theory}
J.~T. Winthrop and C.~R. Worthington, ``Theory of fresnel images. i. plane
  periodic objects in monochromatic light,'' \emph{JOSA}, vol.~55, no.~4, pp.
  373--381, 1965.

\bibitem{tibshirani1996regression}
R.~Tibshirani, ``Regression shrinkage and selection via the lasso,''
  \emph{Journal of the Royal Statistical Society: Series B (Methodological)},
  vol.~58, no.~1, pp. 267--288, 1996.

\bibitem{kim2013specular}
H.~Kim, H.~Jin, S.~Hadap, and I.~Kweon, ``Specular reflection separation using
  dark channel prior,'' in \emph{Proceedings of the IEEE conference on computer
  vision and pattern recognition}, 2013, pp. 1460--1467.

\bibitem{liu2013linearized}
R.~Liu, Z.~Lin, and Z.~Su, ``Linearized alternating direction method with
  parallel splitting and adaptive penalty for separable convex programs in
  machine learning,'' in \emph{Asian Conference on Machine Learning}, 2013, pp.
  116--132.

\bibitem{zhu1997algorithm}
C.~Zhu, R.~H. Byrd, P.~Lu, and J.~Nocedal, ``Algorithm 778: L-bfgs-b: Fortran
  subroutines for large-scale bound-constrained optimization,'' \emph{ACM
  Transactions on Mathematical Software (TOMS)}, vol.~23, no.~4, pp. 550--560,
  1997.

\bibitem{parikh2014proximal}
N.~Parikh, S.~Boyd \emph{et~al.}, ``Proximal algorithms,'' \emph{Foundations
  and Trends{\textregistered} in Optimization}, vol.~1, no.~3, pp. 127--239,
  2014.

\bibitem{wen2019convolutional}
S.~Wen, Y.~Zheng, F.~Lu, and Q.~Zhao, ``Convolutional demosaicing network for
  joint chromatic and polarimetric imagery,'' \emph{Optics letters}, vol.~44,
  no.~22, pp. 5646--5649, 2019.

\bibitem{wang2004image}
Z.~Wang, A.~C. Bovik, H.~R. Sheikh, E.~P. Simoncelli \emph{et~al.}, ``Image
  quality assessment: from error visibility to structural similarity,''
  \emph{IEEE transactions on image processing}, vol.~13, no.~4, pp. 600--612,
  2004.

\bibitem{rahmann2001reconstruction}
S.~Rahmann and N.~Canterakis, ``Reconstruction of specular surfaces using
  polarization imaging,'' in \emph{Proceedings of the 2001 IEEE Computer
  Society Conference on Computer Vision and Pattern Recognition. CVPR 2001},
  vol.~1.\hskip 1em plus 0.5em minus 0.4em\relax IEEE, 2001, pp. I--I.

\bibitem{hedley2005simple}
J.~Hedley, A.~Harborne, and P.~Mumby, ``Simple and robust removal of sun glint
  for mapping shallow-water benthos,'' \emph{International Journal of Remote
  Sensing}, vol.~26, no.~10, pp. 2107--2112, 2005.

\bibitem{liu2020separate}
Y.~Liu and F.~Lu, ``Separate in latent space: Unsupervised single image layer
  separation.'' in \emph{AAAI}, 2020, pp. 11\,661--11\,668.

\bibitem{cao2019makeup}
C.~Cao, F.~Lu, C.~Li, S.~Lin, and X.~Shen, ``Makeup removal via bidirectional
  tunable de-makeup network,'' \emph{IEEE Transactions on Multimedia}, vol.~21,
  no.~11, pp. 2750--2761, 2019.

\end{thebibliography}
	
	%

	\vspace{-40 mm}
	\begin{IEEEbiography}[{\includegraphics[width=1in,height=1.25in,clip,keepaspectratio]{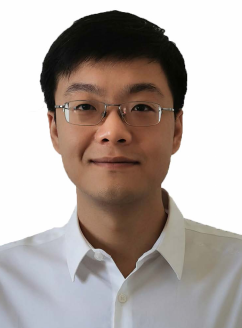}}]{Sijia Wen} is currently pursuing the Ph.D. degree in technology of computer application with the State Key Laboratory of Virtual Reality Technology and Systems, School of Computer Science and Engineering, Beihang University. His research interests include computer vision and polarization-based vision analysis.
	\end{IEEEbiography}
	\vspace{-40 mm}
	\begin{IEEEbiography}[{\includegraphics[width=1in,height=1.25in,clip,keepaspectratio]{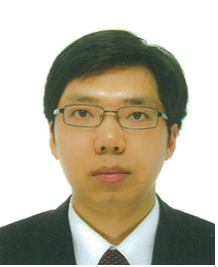}}]{Yinqiang Zheng} (Senior Member, IEEE) received the bachelor’s degree from the Department of Automation, Tianjin University, Tianjin, China, in 2006, the master's degree in engineering from Shanghai Jiao Tong University, Shanghai, China, in 2009, and the Ph.D. degree in engineering from the Department of Mechanical and Control Engineering, Tokyo Institute of Technology, Tokyo, Japan, in 2013. He is currently an Associate Professor with The University of Tokyo, Japan. His research interests include image processing, computer vision, and mathematical optimization.
	\end{IEEEbiography}
	\vspace{-40 mm}
	\begin{IEEEbiography}[{\includegraphics[width=1in,height=1.25in,clip,keepaspectratio]{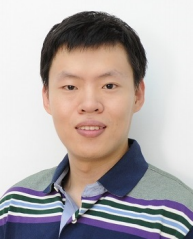}}]{Feng Lu} (Member, IEEE) received the B.S. and M.S. degrees in automation from Tsinghua University in 2007 and 2010, respectively, and the Ph.D. degree in information science and technology from The University of Tokyo in 2013. He is currently a Professor with the State Key Laboratory of Virtual Reality Technology and Systems, School of Computer Science and Engineering, Beihang University. His research interests include computer vision, human–computer interaction, and augmented intelligence. 
	\end{IEEEbiography}
	
	%
	%
	
	
	

\end{document}